\documentclass[lettersize,journal]{IEEEtran}
\usepackage{amsmath,amsfonts}
\usepackage{array}
\usepackage{subfigure}
\usepackage{textcomp}
\usepackage{stfloats}
\usepackage{url}
\usepackage{verbatim}
\usepackage{graphicx}

\def\BibTeX{{\rm B\kern-.05em{\sc i\kern-.025em b}\kern-.08em
    T\kern-.1667em\lower.7ex\hbox{E}\kern-.125emX}}
\usepackage{balance}

\usepackage{booktabs}
\usepackage{multicol}
\usepackage{multirow}

\usepackage{makecell}
\usepackage{threeparttable}
\usepackage{cite}

\usepackage{bm}
\usepackage{color}

\usepackage{dsfont}
\usepackage{eufrak}
\usepackage{graphicx}
\usepackage{float}
\usepackage{placeins}
\usepackage{times}

\usepackage{soul}
\usepackage{url}
\usepackage[utf8]{inputenc}

\usepackage[noend]{algpseudocode}
\usepackage{algorithm}

\newcommand{\cb}{\mathbf{c}}
\newcommand{\wb}{\mathbf{w}}

\newcommand{\bb}{\boldsymbol{\beta}}

\newcommand{\w}{\boldsymbol{w}}
\newcommand{\ba}{\boldsymbol{\alpha}}

\newcommand{\bo}{\boldsymbol{o}}

\newcommand{\namelong}[1]{XXX}
\newcommand{\nameshort}[1]{OStr-DARTS}
\newcommand{\nameshortv}[1]{OStr-DARTS*}

\newtheorem{prop}{Proposition}

\def\all_in_one{all_in_one}
\makeatletter
\newcommand{\printfnsymbol}[1]{%
  \textsuperscript{\@fnsymbol{#1}}%
}
\makeatother

\begin{document}
\title{OStr-DARTS: Differentiable Neural Architecture Search based on Operation Strength}
\author{Le~Yang,
        Ziwei~Zheng,
        Yizeng~Han,
        Shiji~Song, \IEEEmembership{Senior Member, IEEE}, 
        \\ Gao~Huang,
        and~Fan~Li, \IEEEmembership{Senior Member, IEEE}

\thanks{This work was supported in part by the National Natural Science Foundation of China (62206215), in part by the China Post-Doctoral Science Foundation (2022M712537), in part by the China National Post Doctoral Program for Innovative Talents (BX2021241) and in part by CCF-BAIDUOF (2021024).}
\thanks{Le~Yang, Ziwei~Zheng, Fan~Li are with the School of Information and Communications Engineering, Xi’an Jiaotong University, Xi’an, 710049, China. (e-mail: yangle15@xjtu.edu.cn, ziwei.zheng@stu.xjtu.edu.cn, lifan@mail.xjtu.edu.cn.}
\thanks{Yizeng~Han is with the DAMO Academy, Alibaba group, Beijing, 100020, China. (hanyizeng.hyz@alibaba-inc.com) }
\thanks{Shiji~Song, Gao~Huang are with with the Department of Automation, Tsinghua University, Beijing 100084, China (email: shijis@mail.tsinghua.edu.cn, gaohuang@tsinghua.edu.cn)}
}

\markboth{Journal of \LaTeX\ Class Files,~Vol.~18, No.~9, September~2020}%
{How to Use the IEEEtran \LaTeX \ Templates}

\maketitle

\begin{abstract}
Differentiable architecture search (DARTS) has emerged as a promising technique for effective neural architecture search, and it mainly contains two steps to find the high-performance architecture: First, the DARTS supernet that consists of mixed operations will be optimized via gradient descent. Second, the final architecture will be built by the selected operations that contribute the most to the supernet. Although DARTS improves the efficiency of NAS, it suffers from the well-known degeneration issue which can lead to deteriorating architectures. Existing works mainly attribute the degeneration issue to the failure of its supernet optimization, while little attention has been paid to the selection method. In this paper, we cease to apply the widely-used magnitude-based selection method and propose a novel criterion based on operation strength that estimates the importance of an operation by its effect on the final loss. We show that the degeneration issue can be effectively addressed by using the proposed criterion without any modification of supernet optimization, indicating that the magnitude-based selection method can be a critical reason for the instability of DARTS. The experiments on NAS-Bench-201 and DARTS search spaces show the effectiveness of our method.
\end{abstract}


\section{Introduction}\label{sec:introduction}

\begin{figure}
	\centering
	\includegraphics[width=0.48\textwidth]{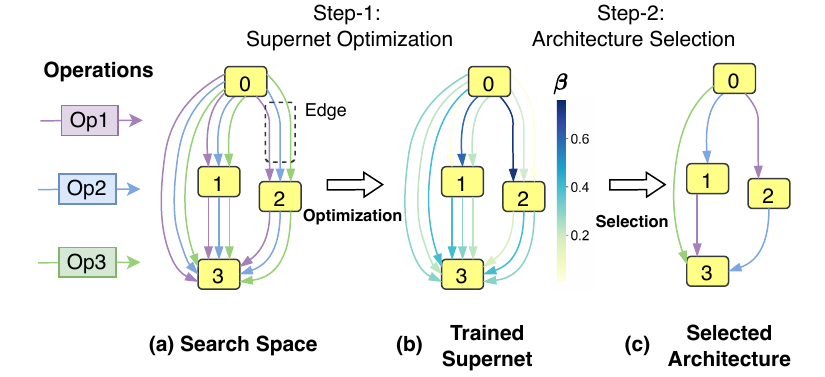}
	\caption{The searching procedure of DARTS: (a) The continuous search space that applies a mixture of candidate operations on each edge. (b) The trained supernet obtained by jointly optimization of $\wb$ and $\ba$. (c) The final architecture selected by $\bb$ ($\bb=softmax(\ba)$). }
	\vspace{-8pt}
	\label{fig_1}
\end{figure}

\IEEEPARstart{I}{n} recent years, neural architecture search (NAS)~\cite{zoph2017neural} has shown its potential in automatically discovering network architectures with high performance. Early works on NAS are mostly realized by reinforcement learning (RL)~\cite{zoph2017neural,zoph2018learning,baker2016designing} and evolutionary algorithms~\cite{real2019regularized,liu2017hierarchical,chen2021modulenet}, which commonly need massive computation overheads, consuming hundreds of GPU days for searching. To improve the search efficiency, researchers propose one-shot methods~\cite{zhang2020one,dong2019one,pham2018efficient} that adopt weight sharing strategy for supernet optimization and then derive the final architecture from the optimized supernet. Based on the idea of one-shot NAS, Liu et al.~\cite{liu2018darts} propose differentiable architecture search (DARTS) that allows architecture parameters to be optimized via a gradient-based algorithm based on continuous relaxation of the architecture representation, which makes architecture search more efficient. The searching procedure of DARTS contains two major steps (shown in Fig.~\ref{fig_1}): 1) The \textit{supernet optimization} step that jointly learns the model weights and architecture parameters via a gradient-based algorithm; 2) The \textit{architecture selection} step that finds the operations with the largest architecture parameters to build final architecture, \textit{a.k.a.} magnitude-based selection method.

Although DARTS enjoys high computational efficiency, it frequently suffers from searching instability~\cite{zela2019understanding}. The performance of the final architecture obtained by the magnitude-based selection method can only be guaranteed when an implicit assumption holds~\cite{wang2021rethinking}: the value of an architecture parameter can correctly reflect the contribution of the corresponding operation to the supernet. However, this assumption cannot be held in many cases. On the one hand, recent researches~\cite{hong2020dropnas,wang2021rethinking} show that the architecture parameters corresponding to the operations with fewer parameters, especially the skip connections, tend to be large after long-epoch weight-sharing optimization. On the other hand, as the searching procedure in DARTS ignores the coupling relationship of the parameters, the architecture parameter of a certain operation cannot be used individually to indicate the importance of this operation~\cite{gu2021dots}. Therefore, the performance of the final architecture obtained by DARTS can collapse when applied with the magnitude-based selection method.

Most existing research proposes to improve the supernet optimization procedure (Step-1 in Fig.~\ref{fig_1}) to address the issue by ensuring the applicability of the magnitude-based selection method in DARTS. Early works mainly stop the search procedure before the performance collapses based on some handcrafted criteria~\cite{liu2018darts,zela2019understanding,liang2019darts+}. Different regularization terms for architecture parameters~\cite{ye2022b} and the norm of Hessian matrix w.r.t $\ba$~\cite{chen2020stabilizing,zela2019understanding} are also developed to stabilize the searching procedure, preventing the domination of skip connections. Moreover, researchers design methods to promote fair optimization for operations during searching~\cite{hong2020dropnas,zheng2021ad}. Other improvements for DARTS~\cite{gu2021dots,xue2021idarts} propose to decouple the interaction between different operations to improve the correlation between the learned architecture parameters and the selected architectures. While all these works still apply the magnitude-based method for generating the final architecture, little attention has been paid to improving the selection process (Step-2 in Fig.~\ref{fig_1}).

Designing a criterion that can accurately reflect the contribution of an operation to the supernet is a challenging problem. The original selection problem leads to an NP-optimization discrete problem, which cannot be effectively solved by most optimization methods. Therefore, the key challenge to designing a good selection criterion is how to effectively find a good solution to the selection problem. In this paper, we propose a novel selection criterion, named \textit{operation strength}, which is derived from approximating the solution of the architecture selection problem by measuring operations' effect on the loss. The operations corresponding to the largest operation strength will be considered important and selected to build the final network after supernet optimization. We call the differentiable neural architecture search based on the proposed operation strength as \nameshort{}. With \nameshort{}, we find that the degeneration problem can be effectively addressed even without any modification of the supernet optimization procedure, indicating that the widely applied magnitude-based architecture selection method can be the main reason for the instability. Our contributions can be summarized as follows:

\begin{itemize}
	\item We develop a novel DARTS method, \nameshort{}, based on the proposed architecture selection method using operation strength, which addresses the degeneration issue by using the proposed selection criterion without any modifications for the supernet optimization.
	\item This work explores another possible way to improve DARTS by designing appropriate criterion for final architecture selection rather than only focusing on the supernet optimization procedure. 
	\item We provide both theoretical and empirical analyses that advocate for operation strength as a better indicator to reflect the contribution to the supernet in DARTS.
	\item Our work can be seamlessly combined with the existing improvements for DARTS supernet to achieve better searching performance.
	
\end{itemize}

The remainder of the paper is organized as follows. Section~\ref{sec:related} introduces the related works. In Section~\ref{sec:method}, we provide the preliminary of DARTS and introduce the new selection criterion and the proposed DARTS method, \nameshort{}. Section~\ref{sec:dis} discusses why the proposed selection criterion can avoid the domination of skip connections and provides some discussion. The experimental results are provided in Section~\ref{sec:exp}. We summarize the limitations of \nameshort{} and some possible future work in Section~\ref{lim}, and then conclude the paper in Section~\ref{sec:conclusion}. Code is available at \url{https://github.com/Ziwei-Zheng/OSen-NAS}.

\section{Related Work}
\label{sec:related}

\subsection{NAS and DARTS}
Recent years have witnessed the rapid development of deep learning, most of which are manually designed\cite{he2016deep,huang2019convolutional,huang2018condensenet,yang2021condensenet,wang2024towards}. To improve the efficiency for designing network architectures, researchers start to design NAS algorithms~\cite{zoph2018learning,liu2017hierarchical} to automatically design neural architectures, mixed-precision-quantized networks~\cite{sun2022fast} and Dendritic neural model~\cite{ji2022competitive}. Moreover, NAS methods have been applied to improve the efficiency of modern CNNs and Vision Transformers~\cite{li2022ds,yang2024evolutionary}. Existing NAS approaches can be divided into three categories: RL-based approaches~\cite{zoph2017neural,zoph2018learning,lyu2021multiobjective}, evolutionary-algorithm-based methods~\cite{liu2021survey,real2019regularized,liu2017hierarchical} and gradient-based approaches~\cite{liu2018darts}. Techniques, like knowledge distillation \cite{dong2023diswot} and language models \cite{chen2024evoprompting}, have also been applied to the NAS field. Also, other works~\cite{cai2020once,guo2023pareto} try to reduce the computational demands of NAS methods to solve the obstacles in real deployments. Recent studies have also achieved NAS using multi-agent~\cite{lopes2024manas} or large language model~\cite{nasir2024llmatic} for searching.

\textbf{Differentiable architecture search (DARTS)}~\cite{liu2018darts} fastens and simplifies the searching procedure by enabling NAS to use gradient descent for search. The recent work in~\cite{heuillet2024efficient} provides a comprehensive survey on DARTS. Recent works further improve DARTS by reducing the memory costs during the searching phase~\cite{xu2021partially,dong2019searching}, and have extended the DARTS for zero-shot learning~\cite{yan2021zeronas} and remote sensing applications~\cite{zhang2021rs}. Although DARTS learns differentiable architecture weights and reduces the search costs, its stability has been challenged because of yielding deteriorating architectures during the search procedure. The final architecture can be dominated by parameter-free operations when training epochs become large. To address this issue, researchers~\cite{zela2019understanding,liang2019darts+,liu2018darts} apply different early stopping strategies to stop the searching before the loss landscape becomes precipitous and prevent the performance collapse of the search process. Explicit~\cite{ye2022b} and implicit~\cite{shu2019understanding,chen2020stabilizing} regularization terms are also introduced to the supernet optimization procedure to improve the stability of DARTS. Moreover, a series of works~\cite{hong2020dropnas,chu2020fair,zheng2021ad} propose to balance the imbalance training procedure for different operations caused by the Matthew Effect, which means that operations with fewer parameters would be trained maturely earlier. The degradation issue is also addressed by decoupling the relationship among different operators\cite{xue2021idarts,gu2021dots,wang2021idarts}, especially skip connection~\cite{chu2021darts-}. However, most of the existing DARTS methods only consider improving the optimization procedure for DARTS supernet, the architecture selection process after optimization has received little attention. 

\subsection{Architecture Selection for DARTS} 
Previous research has shown that: after optimization of the supernet, architecture parameters $\alpha$, cannot accurately reflect the importance of each operation~\cite{wang2021rethinking,zela2019understanding}. Wang \textit{et. al.}, ~\cite{wang2021rethinking} have shown that several failure modes of DARTS can be greatly alleviated with a perturbation-based search method correctly picking the operations with the largest contribution to the supernet. However, the perturbation-based selection method is conducted edge by edge on the optimized supernet, where the importance of an operation is estimated by the accuracy change after fine-tuning the supernet till converges when removing a certain operation, leading to high computational costs. Compared to the method in~\cite{wang2021rethinking}, the operation strength proposed in this paper can be effectively approximated by the gradients of architecture parameters, which are readily available during searching and do not need the cumbersome re-train procedure or hyper-parameter selection. Another line of research lies in building the performance predictor to predict the performance of a target network without training, such as~\cite{xie2023architecture,sun2019surrogate}. Although these methods can achieve good performance, they need the sophisticated design of the prediction model and the encoding scheme of neural networks. Our method can be directly applied to most of the existing DARTS methods to achieve high performance.

Due to the strong relationship between NAS and network pruning, researchers develop architecture selection methods by scoring neural networks based on pruning criteria, such as SNIP~\cite{lee2018snip,wang2020picking,tanaka2020pruning,jorge2021progressive}. E.g., \cite{turner2020blockswap} and \cite{mellor2021neural} perform two types of NAS by scoring candidate networks using different pruning criteria. Moreover, \cite{abdelfattah2020zero} examines pruning-at-initialization criteria towards scoring the target subnetwork. Although these NAS methods can select the sub-architecture with high searching speed, they only achieve fair performance. Also, as their search procedure is based on the ranking of all possible architectures, they can not be applied to large search spaces. Similar to ours, recent work~\cite{zhang2021differentiable} proposes to apply the pruning criteria to estimate the operation importance without training for DARTS. However, considering the differences between the pruning and selection procedure (we will illustrate the differences in the next section), the final architecture obtained by~\cite{zhang2021differentiable} can be still limited. Our experiments in Section~\ref{sec:exp} further demonstrate the importance of considering these differences during estimating the importance of an operation in the DARTS framework.

\vspace{-5pt}
\section{method}
\label{sec:method}

This section first briefly reviews the formation of DARTS. Then we rethink the discrete NAS selection problem and introduce how to approximately solve this problem motivated by the techniques from network pruning, based on which we further introduce the formulation of operation strength and the proposed \nameshort{}.

\begin{figure*}
	\centering
	\includegraphics[width=0.98\textwidth]{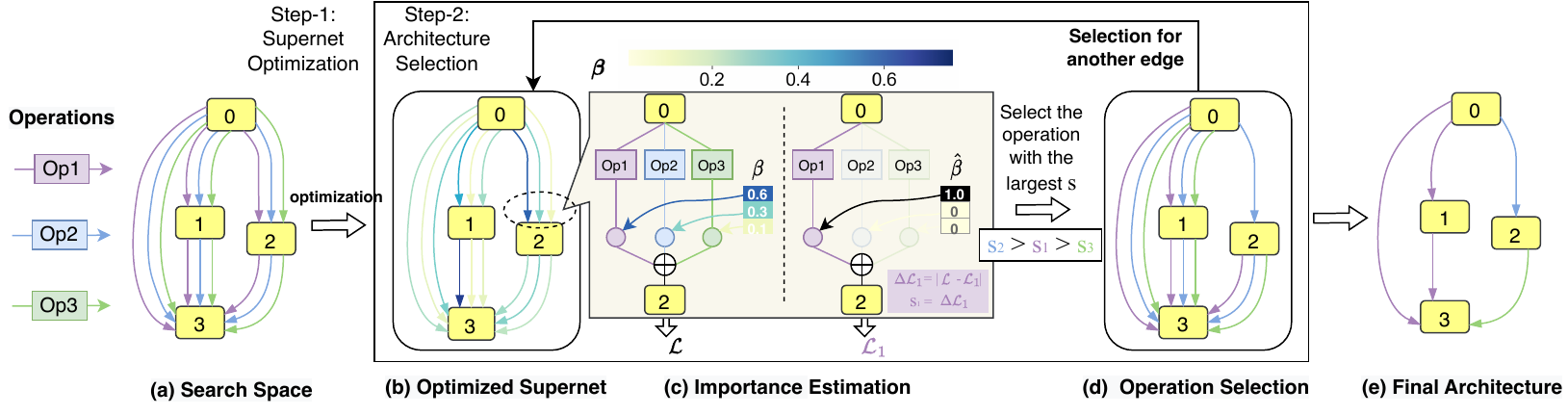}
 \vspace{-10pt}
	\caption{Illustration of the proposed selection criterion (best viewed in color). With the search space (a), we first obtain the optimized supernet (b) via the gradient descent algorithm. Then, importance estimation (c) will be conducted: The Operation strength of Op1 (the purple one) will be calculated by the change of final loss when selecting Op1 as the last operation to replace the mixed one. We do this procedure for each operation and then Op2 is selected as the target operation, although the $\beta_2$ does not have the largest value. We then repeat this procedure for the rest edges to generate the final architecture (e). }
	\label{fig_overview}
\end{figure*}

The overview of our method is shown in Fig.~\ref{fig_overview}. Our work focuses on improving the architecture selection procedure to guarantee the final architecture with high performance can be explored from the optimized supernet. Given a search space, we first obtain the optimized supernet via the gradient descent algorithm. The optimization procedure is not limited to the classical one in DARTS~\cite{liu2018darts}, and can be replaced with any improved methods, such as the memory-efficient PC-DARTS~\cite{xu2021partially}. With the optimized supernet, we conduct the importance estimation on each edge. The importance of an operation is estimated by the \textit{operation strength}, which is calculated by the loss change induced by selecting a specific operation. Then the operation with the largest \textit{operation strength} will be selected on this edge. This procedure is repeated for all edges in the cell to obtain the final architecture. 

\subsection{Formulation of DARTS}
The aim of DARTS~\cite{liu2018darts} is to find the best cell that can stack to form the network with high performance. Each cell is considered as a directed acyclic graph of $N$ nodes, which are latent representations in the network. Moreover, there are $E$ edges in a cell, and an edge $e$ is the connection between different nodes, and consists of a set of candidate operations, $\mathcal{O}$ ($|\mathcal{O}|=P$). To effectively find the optimal operations for each edge, DARTS introduces a mixed operation between every two connected nodes, which can be represented as:
\begin{align}
    \label{eq1}
	\overline{\bo}^e = \overline{o}^e(x_i) = \sum_{o \in \mathcal{O}} \beta_o^e o^e(x_i), \quad \beta_o^e=\frac{\exp (\alpha_o^e)}{\sum_{o'\in\mathcal{O}} \exp (\alpha_{o'}^e) }
\end{align}
where $o^e$ and $\overline{o}^e(x_i)$ mean the operation and the features generated by the mixed operation on edge $e$, and $x_i$ is the input features of edge $e$. Moreover, $\ba\in \mathbb{R}^{P\times E}$ represents the learnable architecture parameters, representing the weights of the corresponding operations. Then the network parameters $\w$ and the architecture parameters $\ba$ can be learned by:
\begin{equation}
    \label{eq2}
    \begin{aligned}
                         \min_{\ba}&\ \ \  \mathcal{L}_{val}(\wb^*(\ba),\ba) \\
        \mathrm{s.t.}  &\ \ \  \wb^*(\ba) = \arg\min_{\wb}\mathcal{L}_{train}(\wb,\ba),
    \end{aligned}
\end{equation}
where network parameters $\w$ and architecture parameters $\ba$ are optimized on the training and validation sets, respectively. At the end of the search phase, the architecture selection step will replace each mixed operation with the most likely operation, $o^e = \arg\max_{o' \in \mathcal{O}}\  \alpha_{o'}^e$, to generate the final target architecture. However, a larger value of $\alpha$ does not necessarily mean that the corresponding operation has more contribution to the whole supernet. Therefore, it is necessary to design a better criterion to accurately estimate the importance of different operations.

\subsection{Rethinking selection problem of DARTS}


Actually, the final selection problem of DARTS can be denoted as\footnote{Here, we consider the one-cell situation for simplicity.}:
\begin{equation}
    \label{a_eq1}
    \begin{aligned}
        \min_{\cb}&\ \ \  \mathcal{L}_{val}( \cb \odot \boldsymbol{O}(\wb^*) ; \mathcal{D}_{val} ) \\
        \mathrm{s.t.}  &\ \ \  \wb^*(\ba) = \arg\min_{\wb} \ \mathcal{L}_{train}(\wb; \ba, \mathcal{D}_{train}), \\
        & \cb \in \{0,1\}^{P\times E}, \ \  \parallel  \!\cb_e\! \parallel_0 \ =  1,  \ \ e = 1,...,E, 
    \end{aligned}
\end{equation}
where $\cb$ represents operation indicators, $\mathcal{D}$ represents the datasets. Moreover, $\boldsymbol{O}$ is the collection of generated features for all operations in the cell. We use $\cb \odot \boldsymbol{O}$ to denote the obtained features masked by the learned indicator matrix $\cb$. The constraint, $\parallel\! \cb_e \!\parallel_0 =1$, forces that each edge will retain the best operation to build the final cell. Due to the complexity of solving the discrete optimization problem (\ref{a_eq1}), DARTS relaxes $\cb$ to the continuous $\bb \in[0,1]^{P\times E}$ for easier optimization.

While, in network pruning fields, a similar discrete optimization problem is solved by exploring the weight connection sensitivity of the loss function as a proxy~\cite{lee2018snip}. The pruning problem can be represented as:
\begin{align}
    \label{a_eq2}
    \min_{\cb, \wb} \quad & \mathcal{L} (\cb \odot  \wb; \mathcal{D}) \\
    s.t.\quad  & \cb \in \{0,1\}^m,  \parallel \! \cb \! \parallel_0 \ \leq k. \nonumber 
\end{align}
where $m$ is the number of total network parameters, and $k$ represents the sparsity level. To solve the problem with high efficiency, pruning methods circumvent this problem by finding the weights with the maximal impacts on the final objective function. The similarities between the problem (\ref{a_eq1}) and (\ref{a_eq2}) motivate us to find the operations that affect most on the loss function in the final selection problem. 

\begin{figure}
	\centering
 \vspace{-10pt}
	\includegraphics[width=0.4\textwidth]{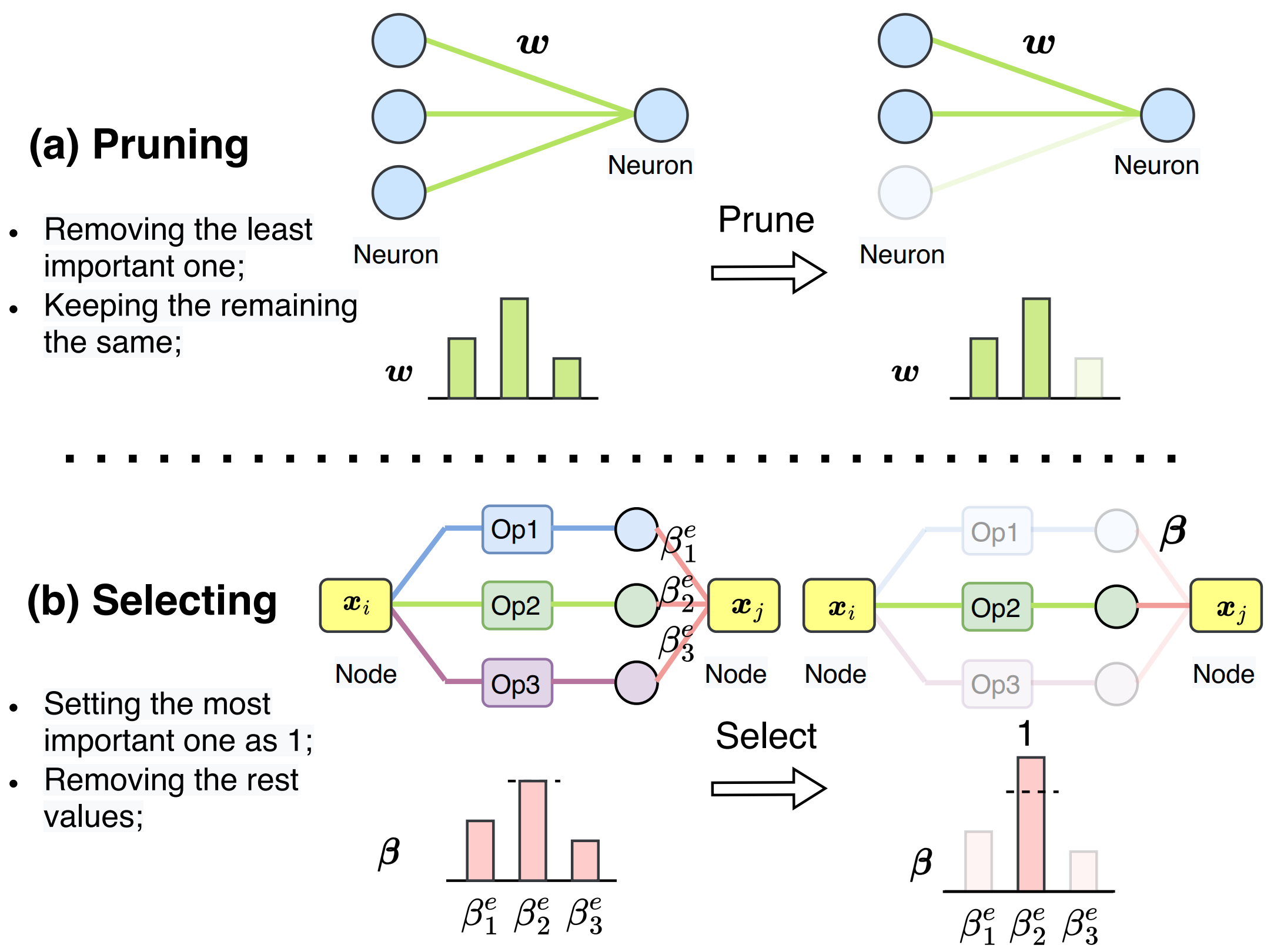}
	\caption{The differences between (a) pruning and (b) architecture selection. }
	\label{fig_2}
 \vspace{-10pt}
\end{figure}
However, importance estimation for a neuron (or a kernel and a filter) in ~\cite{lecun1989optimal,molchanov2019importance,molchanov2016pruning} is significantly different from that for an operation. As is shown in Fig.~\ref{fig_2} (a), network pruning aims at removing the unimportant weights and keeping the rest of them. While, architecture selection needs to select the most important operation and set the $\bb^e$ as $\hat{\bb}_o^e$, where $\hat{\bb}^e_o \in \mathbb{R}^{P\times 1}$ is the all-zero vector except $\beta_o^e=1$ (Fig.~\ref{fig_2} (b)). Such a difference can lead to the sub-optimal when naively applying the importance estimation strategy from pruning in architecture selection, which measures the loss change induced by removing an operation.

\begin{figure}
 \vspace{-10pt}
	\centering
	\includegraphics[width=0.43\textwidth]{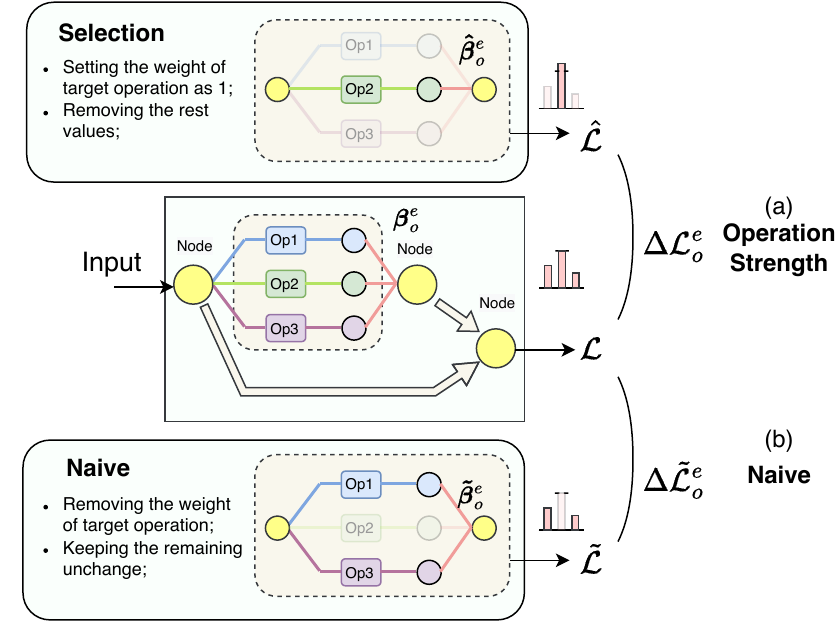}
	\caption{Importance estimation between (a) our method and (b) the naive implementation as it in network pruning.}
	\label{fig_3}
 \vspace{-15pt}
\end{figure}

A more reasonable importance estimation method for architecture selection should consider the effect induced by selecting the target operation, which is shown in Fig. \ref{fig_3}~(a). Suppose that $\overline{\bo}^e$ is the generated feature of the mixed operation on edge $e$, and $\bo_o^e$ is the feature from operation $o$, the importance of the operation $o$ on edge $e$ is defined as the change in loss ($\Delta\mathcal{L}^e_o$) when this operation is selected on edge $e$, resulting in using $\bo^e_o$ to substitute the mixed features. Following~\cite{lee2018snip,molchanov2016pruning}, the effect of selecting the target operation can be measured by:
\begin{align}
    \label{eq4}
	\Delta\mathcal{L}^e_o = | \mathcal{L}(\overline{\bo}^e; \mathcal{D}) - \mathcal{L}(\bo^e_o; \mathcal{D})|.
\end{align}
The loss change can be approximated by the first-order Taylor expansion\footnote{We omit the $\mathcal{D}$ for simplicity}:
\begin{align}
    \label{eq5}
    \begin{aligned}
\Delta\mathcal{L}^e_o  & = |\frac{\partial \mathcal{L}}{\partial \bo^e_o} \  (\bo^e_o - \overline{\bo}^e) + R_1(\bo^e_o - \overline{\bo}^e)   |\\
    &\approx |\frac{\partial \mathcal{L}}{\partial \overline{\bo}^e} \  \frac{\partial \overline{\bo}^e}{\partial \bo^e_o}\  (\bo^e_o - \overline{\bo}^e)| =  \beta^e_o \ | \frac{\partial \mathcal{L}}{\partial \overline{\bo}^e} (\bo^e_o-\overline{\bo}^e)|,
    \end{aligned}
\end{align}
where $R_1(\bo^e_o - \overline{\bo}^e)$ is the Lagrange form of the remainder, which can be omitted in the approximation and equals to:
\begin{align}
    \label{eq_remainder}
    \begin{aligned}
    R_1(\bo^e_o - \overline{\bo}^e) = \frac{1}{2} \  \frac{\partial^2 \mathcal{L}}{ { \partial \bo^e_o} ^2} |_{\bo^e_o = \xi}\  (\bo^e_o-\overline{\bo}^e)^2,
    \end{aligned}
\end{align}
for some vector $\xi$ between $\bo^e_o$ and $\overline{\bo}^e$.

\newtheorem{remark}{Remark}
\begin{remark}
\label{fn1} \textit{(R1)} The dimensions of variables in Eq.(\ref{eq5}) and (\ref{eq_remainder}) will be vectorized (or reshaped) to the appropriate dimensions to conduct matrix multiplication. E.g., in Eq.(\ref{eq5}) suppose $\frac{\partial \mathcal{L}}{\partial \bo^e_o} \in \mathbb{R}^{c\times h \times w}$, where $c$, $h$, $w$ are feature channels and resolutions, then we vectorize it to $\mathbb{R}^{1 \times chw}$. Similarly, $(\bo^e_o - \overline{\bo}^e)\in \mathbb{R}^{c\times h \times w}$ will be vectorized to the size of $\mathbb{R}^{chw \times 1}$. Also for other similar formulations in this paper. We omit this reshaping or vectorization procedure for better visualization of the formulations.
\end{remark}

\vspace{-10pt}
\subsection{Operation Strength}
The \textit{operation strength} of an operation $o$ is defined by the absolute change of final loss when the optimized supernet selects this operation as the final operation to replace the mixed one, which can be represented by$^{\text{R}\ref{fn1}}$:
\begin{equation}
    \label{eq3}
   s^e_o = \Delta\mathcal{L}^e_o =\beta^e_o \ | \frac{\partial \mathcal{L}}{\partial \overline{\bo}^e} (\bo^e_o-\overline{\bo}^e)|.
\end{equation}

Note that the gradient of architecture parameters $\ba$ of the supernet can be written as$^{\text{R}\ref{fn1}}$
\begin{align}
	\label{eq9}
    \begin{aligned}
	\frac{\partial \mathcal{L}} { \partial \alpha_o^e}  =g_o^e &= \frac{\partial \mathcal{L}} { \partial  \overline{\bo}^e } \  \frac{\partial \overline{\bo}^e}{\partial \bb}  \frac{\partial \bb} { \partial  \beta_o^e} (\frac{\partial \beta_o^e}{\partial \alpha_o^e} + \sum_{o' \neq o} \frac{\partial \beta_{o'}^e}{\partial \alpha_o^e}) \\
    &= \frac{\partial \mathcal{L}} { \partial  \overline{\bo}^e } \  \frac{\partial \overline{\bo}^e}{\partial \bb}  \frac{\partial \bb} { \partial  \beta_o^e}  (\beta_o^e(1-\beta_o^e) - \sum_{o' \neq o} \beta_o^e \beta_{o'}^e) \\
    & =\beta_o^e \cdot \frac{\partial \mathcal{L}} { \partial  \overline{\bo}^e } ( \bo^e_o -\overline{\bo}^e),
    \end{aligned}
\end{align}
then we have $s_o^e = |g_o^e|$. For $L$-cells case, we still have$^{\text{R}\ref{fn1}}$:
\begin{align}
	\label{eq10}
    \begin{aligned}
	s_o^e & = |g_o^e| = |\frac{\partial \mathcal{L}} { \partial \alpha_o^e}| \\
    & =\beta_o^e \cdot \sum_{l=1}^L  \frac{\partial \mathcal{L}} { \partial  \overline{\bo}^l_e } ( \bo_e^{ol} -\overline{\bo}^{el}),
    \end{aligned}
\end{align}
where $\overline{\bo}^{el}$ is the generated feature of the mixed operation on edge $e$ at $l$-th cell, and $\bo_e^{ol}$ is the feature from operation $o$ at $l$-th cell. Therefore, the proposed operation strength can be obtained directly from the architecture parameter gradients without any computational cost, which is readily available during the optimization phase.

During the searching phase, the gradients of architecture parameters will be computed using a minibatch of data, and the final $s_o^e$ will be calculated for $B$ minibatches, and averaged over $B$. However, after sufficient training of supernet, the gradient $\frac{\partial \mathcal{L}}{\partial \ba}$ should be 0 at the optimal ideally, which means that: $\frac{\partial \mathcal{L}}{\partial \ba} \rightarrow  0$, and therefore, the operation strength will tend to be zero, namely, $\mathbb{E}_{\mathcal{D}}(s_o^e)\rightarrow 0 $. Empirically, the gradient w.r.t $\ba$ can be represented as $\frac{1}{N} \sum_n \frac{\partial \mathcal{L}_n}{\partial \ba} \approx \boldsymbol{0}$, where $\mathcal{L}_n$ denotes the loss of $n$-th sample. Nevertheless, as a result of stochastic gradient evaluations, the practical operation strength would never be zero. The operation strength takes the absolute value of the gradient and is averaged over $B$ different minibatches. This means that $s_o^e$ is actually computed by:
\begin{align}
    \label{eq11}
	s_o^e =\frac{1}{B} \sum_{b=1}^B | (g_o^e)_b| ,
\end{align}
where $(\cdot)_b$ means the results for $b$-th minibatch. As we can see, in the extreme case that the batch size equals 1, the operation strength will tend to $\mathbb{E}_{\mathcal{D}}(|\frac{\partial \mathcal{L}}{\partial \ba}|)$, which satisfies $\mathbb{E}_{\mathcal{D}}(|\frac{\partial \mathcal{L}}{\partial \ba}|)\! \propto\! \sigma$ ($\sigma$ is the standard deviation of $\frac{\partial \mathcal{L}}{\partial \ba}$) under the assumption that samples are drawn with \textit{i.i.d}. Therefore, the calculated operation strength is proportional to the standard deviation of $\frac{\partial \mathcal{L}}{\partial \ba}$, a value which is empirically more informative as it is stated in~\cite{molchanov2019importance}. 

With the calculated operation strength for each operation, we can build the final architecture with the operation corresponding to the largest operation strength on each edge. 

\vspace{-15pt}
\subsection{\nameshort{}}
\begin{algorithm}
    \caption{$\ $\nameshort{}}
    \label{al1}
    \renewcommand{\algorithmicrequire}{\textbf{Input:}}
    \renewcommand{\algorithmicensure}{\textbf{Output:}}
    \begin{algorithmic}[1]
    \Require $A_0$; $C$; $T$; $Er$;\\
	$t=1$;$cnt=0$;
    \While{$cnt < C$ or $t\leq T$} 
        \State Update network weights $\wb$ and parameters $\ba$
        \If{$Er$ or $t==T$}  
		\State Calculate $s^e_o$ for each operation $o$ in each edge $e$
        \State Select the architecture $A_t$ based on each $s^e_o$
		\If{$A_{t-1}==A_t$}  \State $cnt = cnt + 1$
        \Else
        \State $cnt = 0$
        \EndIf
		\EndIf
        \State $t = t + 1$
    \EndWhile
    \Ensure Final selected architecture $A^* = A_t$
    \end{algorithmic}
\end{algorithm}
We now introduce the proposed differentiable architecture search based on operation strength, \nameshort{}. The optimization procedure can be either the same as the classical bi-level method in~\cite{liu2018darts}, or be other improved ones, such as the memory-efficient procedure in~\cite{xu2021partially}. During each searching step, we can directly obtain the $s^o_e$ for different operations on each edge without extra costs since the gradient information has already been calculated during the optimization procedure. Moreover, as the resulted architectures of \nameshort{} converge fast in some search spaces, we can stop the searching procedure early if the selected architecture does not change for a given number of epochs $C$. We use $Er$ to define if we want an early stop criterion during the optimization and use $T$ and $A_0$ to denote the total number of searching epochs and the initial random architecture. The fast convergence speed of the final further demonstrates the stability of our method. The complete algorithm is given in Algorithm \ref{al1}.

\section{Theoretical Analysis and Discussions of OStr-DARTS}
\label{sec:dis}

\subsection{Avoid the domination of skip connections}
\begin{figure}[htp]
	\centering
	\includegraphics[width=0.48\textwidth]{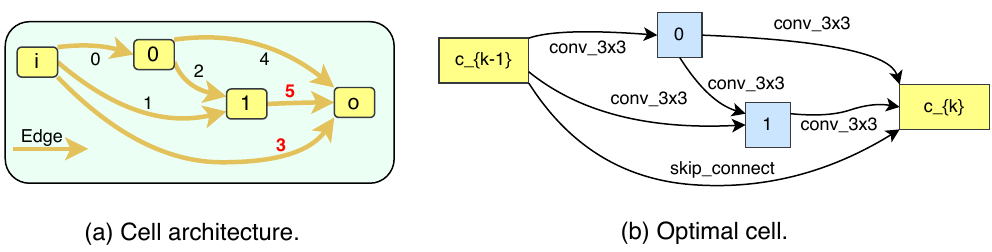}
    \caption{NAS-Bench-201 search space. (a) The cell architecture. (b) The optimal cell.}
	\label{fig_cell}
 \vspace{-10pt}
\end{figure}

We show why OStr-DARTS can avoid the domination of skip connections. Following~\cite{wang2021rethinking}, we provide a case study about the skip connection in DARTS and show that the optimal solution of $\beta_{skip}$ in the sense of minimizing the variance of feature map estimation. 
\begin{prop}
\textit{Without loss of generality, consider one cell from a simplified search space consisting of two operations: skip connection and conv. Let $m^*$ denote the optimal feature map, which is shared across all edges according to the unrolled estimation view~\cite{greff2016highway}. Let $o_{conv}(x_e)$ be the output of the convolution operation, and let $x_e$ be the input. Assume $m^*$, $o_{conv}(x_e)$ and $x_e$ are normalized to the same scale. The current estimation of $m^*$ can then be written as:}
\begin{align}
	\overline{\bo}^e = \beta_{skip}x_e +  \beta_{conv} o_{conv}(x_e).\nonumber
\end{align}
\textit{Then optimal $\beta_{skip}$ and $\beta_{conv}$ minimizing $var(\overline{\bo}^e-m^*)$ are:}
\begin{align}
 \beta_{conv} \propto var(x_e-m^*), \ \  \beta_{skip} \propto var(o_{conv}(x_e)-m^*). \nonumber
\end{align}
\end{prop}
The detailed proof can be found in~\cite{wang2021rethinking}. 

From the conclusion from~\cite{wang2021rethinking}, we can see that the magnitudes of $\beta_{skip}$ and $\beta_{conv}$ will converge to the values which one of the residual features is closer to $m^*$ in variance. $x_e$ comes from the mixed output of the previous edge. Since the goal of every edge is to estimate $m^*$ (unrolled estimation), $x_e$ is also directly estimating $m^*$. However, $o_{conv}(x_e)$ is the output of a single convolution operation, so it can deviate from $m^*$ even at convergence. Therefore, in a well-optimized supernet, $x_e$ will naturally be closer to $m^*$ than $o_{conv}(x_e)$, causing $\beta_{skip}$ to be greater, which leads to the domination of skip connections. 

However, $x_e$ being closer to $m^*$ can lead to a lower value of $|\overline{\bo}^e-x_e|$, resulting in the decrease of the operation strength of skip connections. These two terms balance and avoid the domination of skip connections in \nameshort{}.

\begin{figure}[t]
	\centering
	\subfigure[Norms of RF., $\beta$ on Edge-3.]{
	\includegraphics[width=0.23\textwidth]{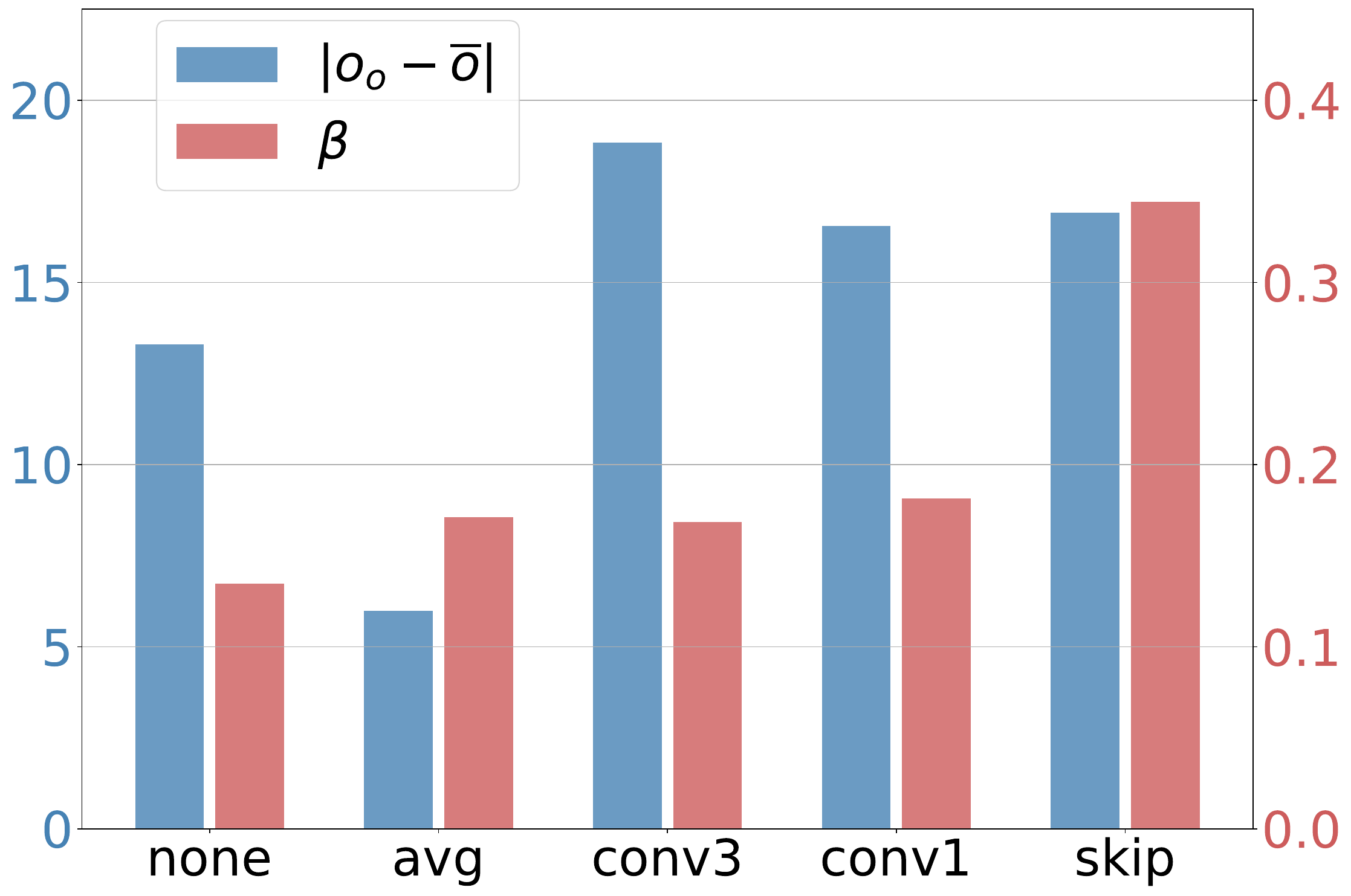}
	}	
	\subfigure[Operation strength on Edge-3.]{
	\includegraphics[width=0.22\textwidth]{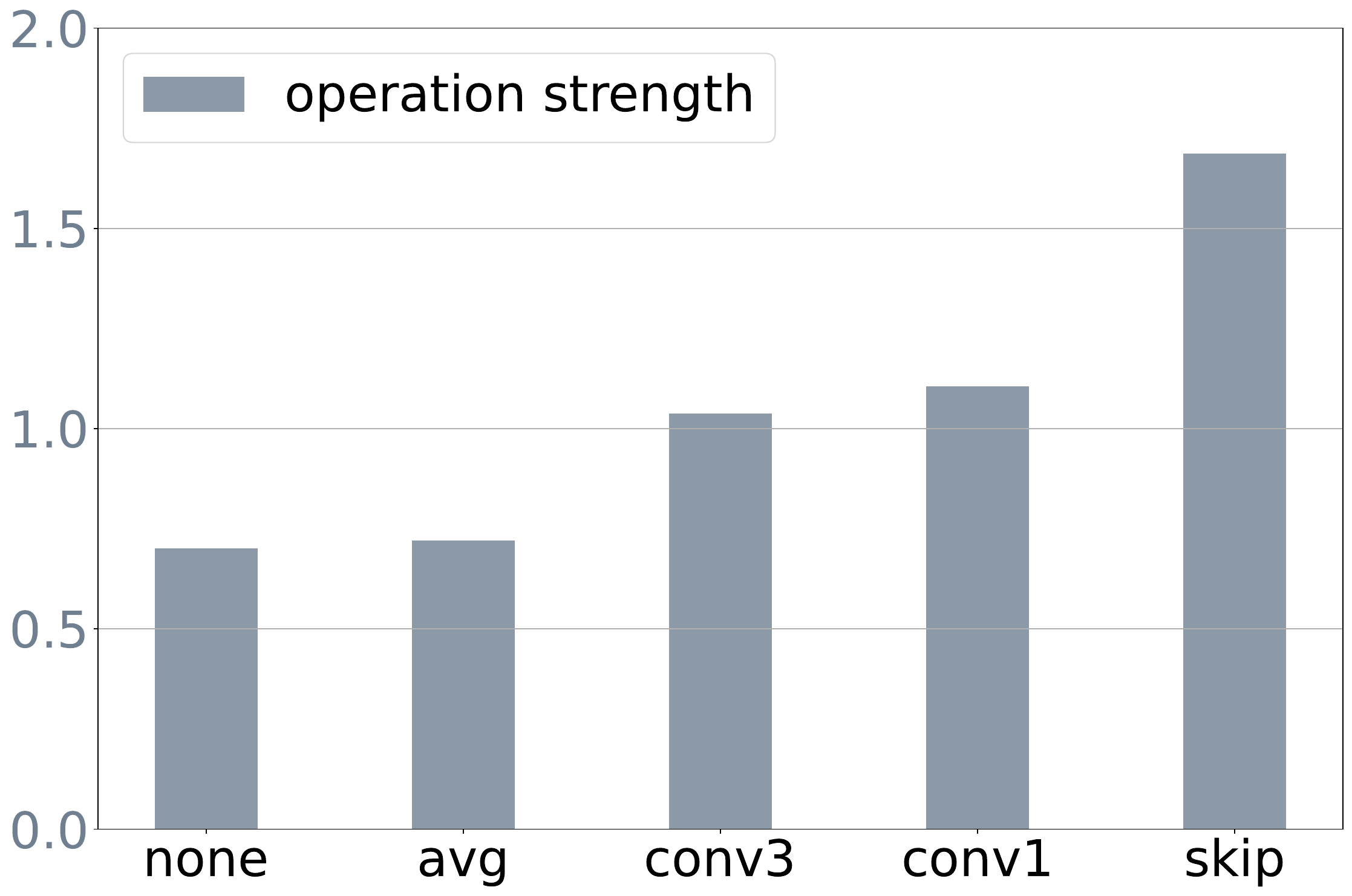}
	}
	\subfigure[Norms of RF., $\beta$ on Edge-5.]{
	\includegraphics[width=0.23\textwidth]{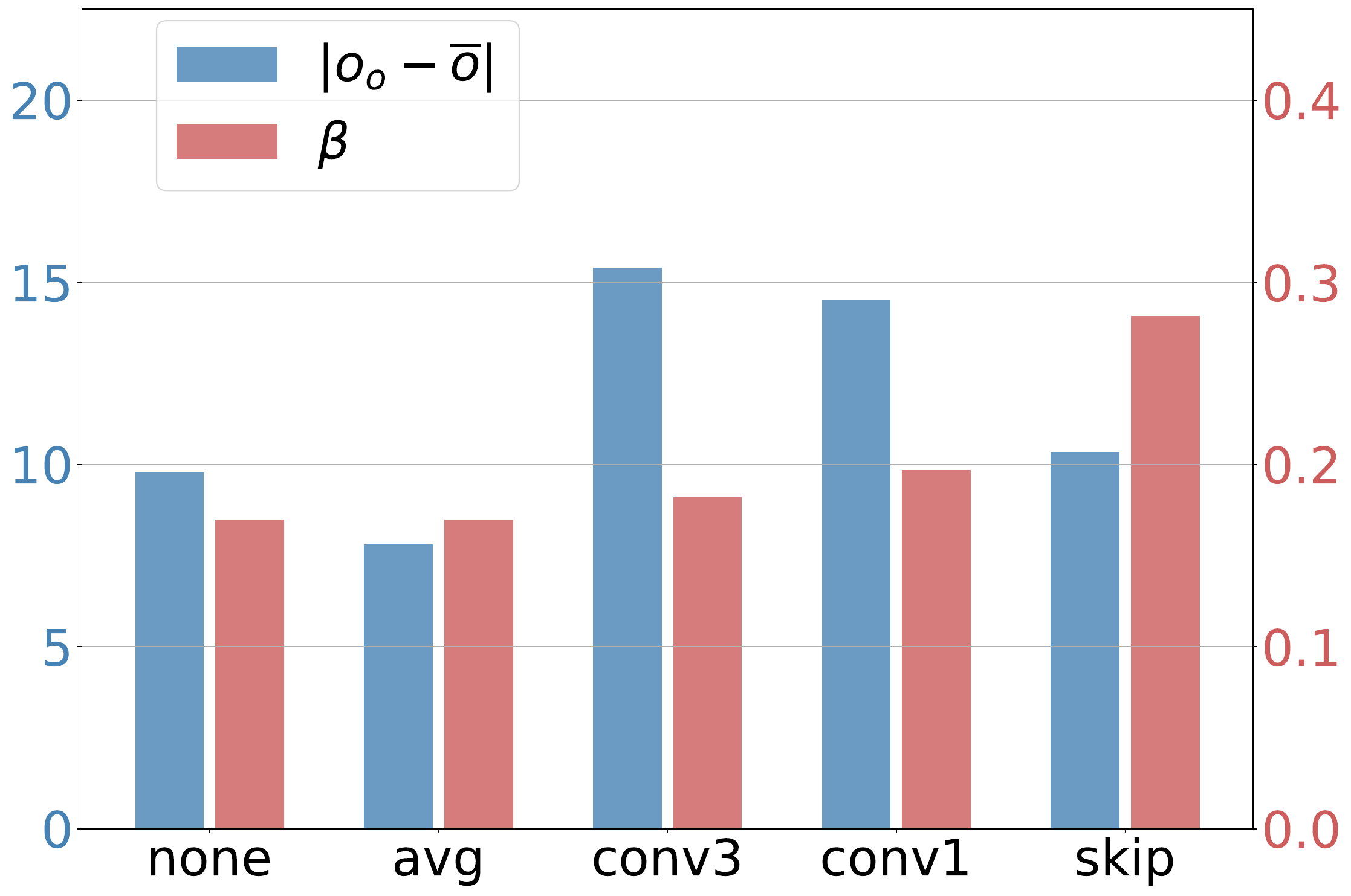}
	}
	\subfigure[Operation strength on Edge-5.]{
	\includegraphics[width=0.22\textwidth]{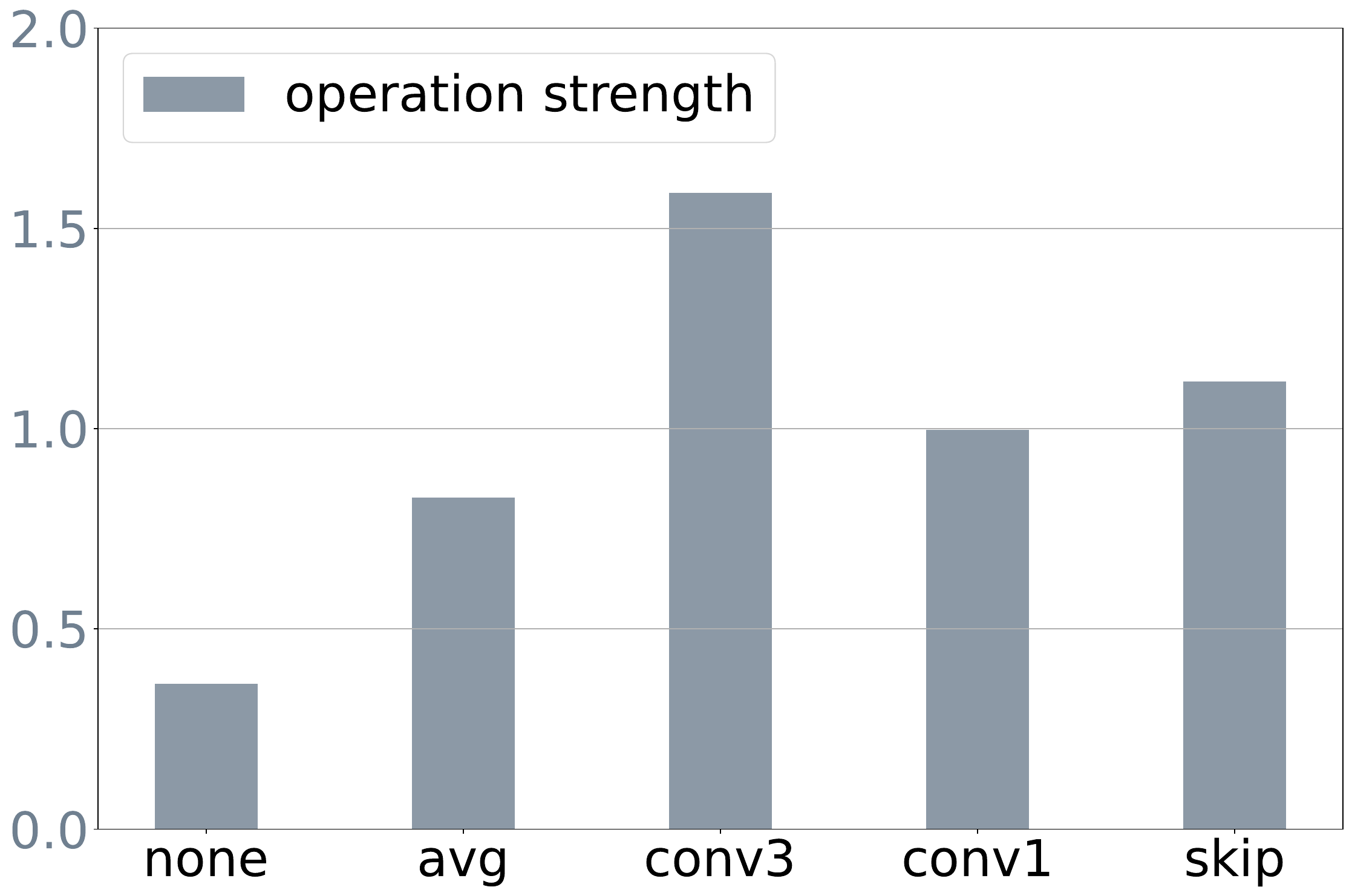}
	}
	\caption{Norms residual features (RF.), $\beta$ and operation strength of different operations on Edge-3 and Edge-5.}
	\label{fig_skp}
  \vspace{-10pt}
\end{figure}

Empirical results are also provided in Fig.~\ref{fig_cell} (a), showing the illustration of a cell architecture and the optimal cell\footnote{Here, we mean the relatively optimal, since the optimal cell architectures for the three datasets in NAS-Bench-201 are different.} in NAS-Bench-201 space. The proposed \textit{operation strength} consists of $\bb$ and the norm of $|\bo_{o}^e - \overline{\bo}^e|$, which can be referred to as the residual features (RF). The results in Fig.~\ref{fig_skp} (a) and (c) show that the $\beta$ of skip connection is larger than others, resulting in the domination of skip connections using the magnitude-based selection methods. However, since that operation strength further uses the RF to estimate the importance of an operation, the skip connection with little effect on final loss will be ignored during selection, guaranteeing criterion can select the essential operations from the supernet.

We further investigate what a higher value of the RF norm represents here. Consider the RF as $|\bo_{o}^e - \overline{\bo}^e|$, it can be rewritten as:
\begin{align}
 |\bo_{o}^e - \overline{\bo}^e| &= | \bo_{o}^e \sum_{o'} \beta^e_{o'}  - \sum_{o'} \beta^e_{o'} \bo_{o'}^e   |\\
&=|  \sum_{o' \neq o}  \beta^e_{o'}(\bo_{o}^e  - \bo_{o'}^e ) | \leq \sum_{o' \neq o}  \beta^e_{o'} |\bo_{o}^e  - \bo_{o'}^e  |.
\end{align}
From the inequality, we see that a higher value of RF means that the features $\bo_{o}^e$ obtained by the target operation are distinctive compared to the features $\bo_{o'}^e$ generated by the rest operations on average with the scaled factor $\beta^e_{o'}$. This indicates that the impact of the generated features is also an important factor in selecting the operation contributing to the supernet most, which meets the recent findings in large language model (LLM) fields~\cite{sun2023simple}.

Therefore, operation strength will evaluate the importance of an operation from two aspects: 1) Whether the magnitude of the architecture parameter for the corresponding operation is large. 2) Whether the features generated by this operation are the most distinctive ones. Considering that appropriate applications of skip connections can alleviate the gradient vanishing problem and benefit the performance of modern deep networks~\cite{he2016deep}, the optimal solution of the problem (\ref{a_eq1}) should keep the important skip connections, and eliminate these redundant ones. By introducing the effect of skip connection on the final loss, these important skip connections will be explored and selected to build the final architecture and the optimal cell architecture will be selected (shown in Fig.~\ref{fig_cell} (b)).

\subsection{Operation strength: architecture parameter perspective}
In this subsection, we provide the analysis about why using $|\frac{\partial \mathcal{L}}{\partial \ba}|$ as a selection criterion in Eq. (\ref{eq10}) provides a better final architecture. Again, considering the discrete NAS problem in (\ref{a_eq1}), we may be able to determine the importance of a given operation by measuring its effect on the loss function when the operation is selected, which is shown to be an effective way to solve the discrete problem in network pruning. To measure the effect of a target operation on the loss, one can try to measure the difference in loss with the $\ba$ and $\ba_{o\rightarrow\infty}$:
\begin{align}
 \Delta \mathcal{L}_o  =  |\mathcal{L} (\ba,\wb; \mathcal{D}) - \mathcal{L} (\ba - (\ba-\ba_{o\rightarrow\infty}),\wb; \mathcal{D})|,
\end{align}
where $\ba_{o\rightarrow\infty}$ is the architecture parameter that element $o$ equals to $+\infty$ and others remain the same as $\ba$, resulting in that $\text{Softmax}(\ba_{o\rightarrow\infty}) \rightarrow \hat{\bb}^e_o$. It is easy to see that $\Delta \mathcal{L}_o$ attempts to measure the influence when using operation $o$ on the loss function. Following~\cite{lee2018snip}, $\Delta \mathcal{L}_o$ can be approximated by the directional derivative of $\mathcal{L}$, which can be written as:
\begin{align}
 \Delta \mathcal{L}_o \approx |\lim_{\delta \rightarrow 0} \frac{ \mathcal{L} (\ba,\wb; \mathcal{D}) - \mathcal{L} (\ba - \delta(\ba-\ba_{o\rightarrow\infty}),\wb; \mathcal{D}) }{\delta}|,
\end{align}
where $\delta$ is infinitesimal of higher order. Moreover, note that the direction of $\ba-\ba_{o\rightarrow\infty}$ is the same as the $\hat{\ba}^e_o$ (the all-zero $\ba$ expect for $o$ element equaling to 1), the above equation can be rewritten as:
\begin{align}
 \Delta \mathcal{L}_o \approx |\lim_{\delta \rightarrow 0} \frac{ \mathcal{L} (\ba,\wb; \mathcal{D}) - \mathcal{L} (\ba - \delta\hat{\ba}^e_o,\wb; \mathcal{D}) }{\delta}| = |\frac{\partial \mathcal{L}}{\partial \alpha^e_o}|.
\end{align}
Therefore, operation strength actually estimates the importance of a given operation when the edge selects it as the target one, where the importance is used to approximate the final optimal solution of the original discrete NAS selection problem in (\ref{a_eq1}).

 \vspace{-10pt}
\subsection{The novelty and superiority of operation strength}
In this section, we provide a discussion about why the proposed operation strength criterion works well compared to magnitude-based selection. First, due to the complexity of the DARTS problem, using the magnitude-based method to transform the relaxed continuous optimal solution to the discrete one may not be a good way to solve the original selection problem in (\ref{a_eq1}). As shown in previous researches~\cite{chu2021darts-,ye2022b}, the issue-causing discrete step can be the culprit of the instability of DARTS. While the proposed operation strength ties to directly estimate the optimal solution of the problem (\ref{a_eq1}), avoiding the issue-causing discrete step, and therefore can be more stable when selecting the final architecture. Second, the analysis at the beginning of Section III and Section IV-B both indicate that the operation strength can be used as a good approximation for the optimal solution for the problem (\ref{a_eq1}). Last but not least, the proposed Ostr-DARTS can effectively select the most important skip connections and eliminate the redundant ones. This also guarantees the performance of the final selected architecture using operation strength.

Although both our method and previous work in~\cite{lee2018snip} use the Taylor expansion-based to achieve the importance estimation, they are designed for different purposes. The importance indicator in~\cite{lee2018snip} is specialized for the pruning problem, while we design a new Taylor expansion-based importance indicator for architecture selection in DARTS. This design does not follow the design routine in existing works that all proposed to estimate the importance by removing the target weights/kernels/filters/operations. From Fig.~\ref{fig_2}, we see that pruning and selection procedures are different. Therefore, directly using the existing importance indicators~\cite{lee2018snip} is not reasonable and should be rebuilt and redesigned for the architecture selection in the NAS problem, which motivated this work. Moreover, the experimental results in Section~\ref{sec:exp} further confirm that directly using the importance indicator in~\cite{lee2018snip} can lead to sub-optimal performance, demonstrating the effectiveness and novelty of our method.

\subsection{\nameshortv{}}
Actually, the $\Delta \mathcal{L}_o$ can be calculated using the first-order Taylor expansion with two different formulations expanded at $\bo^e_o$ or $\overline{\bo}^e$, respectively$^{\text{R}\ref{fn1}}$:
\begin{align}
    \label{a_eq111}
    \begin{aligned}
    \Delta\mathcal{L}^e_o &= |\mathcal{L}(\overline{\bo}^e; \mathcal{D}) - \mathcal{L}(\bo^e_o; \mathcal{D})| \approx |\frac{\partial \mathcal{L}}{\partial \bo^e_o} \  (\bo^e_o - \overline{\bo}^e) | \\
    & = \beta^e_o \ | \frac{\partial \mathcal{L}}{\partial \overline{\bo}^e} (\bo^e_o-\overline{\bo}^e)|,
    \end{aligned}
\end{align}
or,
\begin{align}
    \label{a_eq112}
    \begin{aligned}
    \Delta\mathcal{L}^e_o = | \mathcal{L}(\bo^e_o; \mathcal{D}) - \mathcal{L}(\overline{\bo}^e; \mathcal{D})| \approx |\frac{\partial \mathcal{L}}{\partial \overline{\bo}^e} \  (\overline{\bo}^e - \bo^e_o ) |.
    \end{aligned}
\end{align}
We refer to the new selection criterion in Eq. (\ref{a_eq112}) as \textit{operation strength}*, ${s^*}^e_o$, and based on this we can develop the \nameshortv{}. According to Eq.(\ref{a_eq111}), the new indicator can be calculated by ${s^*}^e_o = s_o^e/\beta_o^e$. Our experiments further confirm that \nameshortv{} also boosts the performance compared to the Naive one, indicating the limitation of the direct application of pruning techniques in architecture selection.

\begin{figure}
	\centering
	\includegraphics[width=0.45\textwidth]{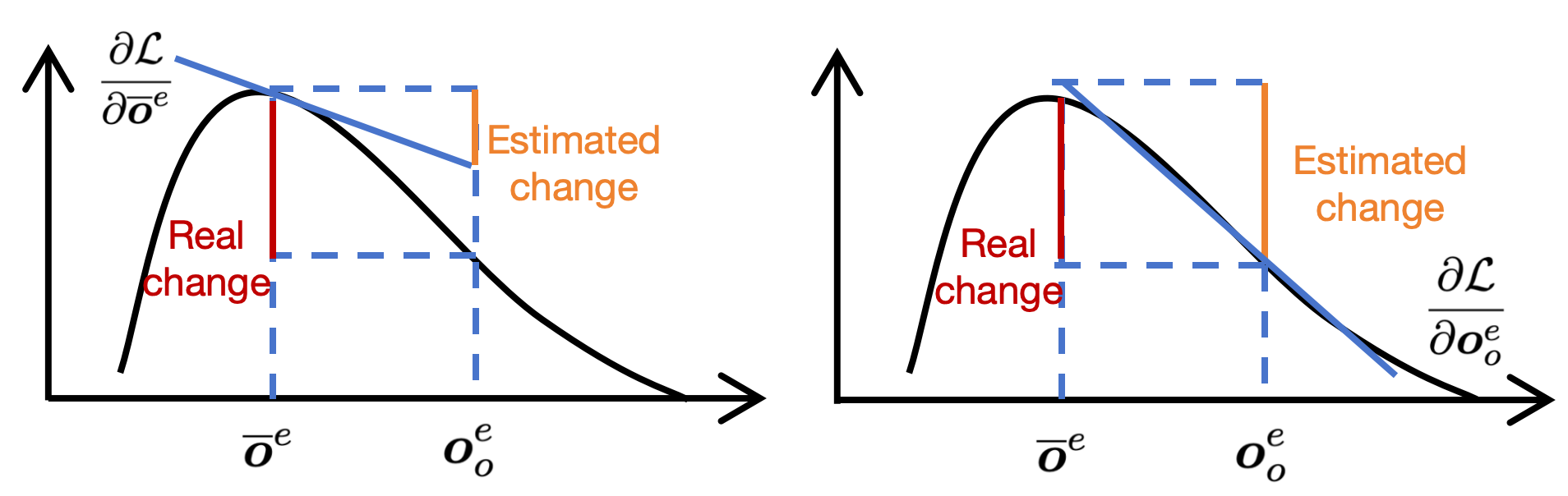}
	\caption{Estimation errors using two different Taylor expansion.}
	\label{fig_toy}
  \vspace{-10pt}
\end{figure}

The difference between \nameshort{} and \nameshortv{} is the expansion points using Taylor expansion. After optimization, the $\overline{\bo}^e$ will tend to be near the optimal point where the gradient can be flattened, resulting in a less accurate estimation of the loss change (illustrated in Fig.~\ref{fig_toy}). While, we find that both \nameshort{} and \nameshortv{} can achieve promising performance in both NAS-Bench-201 and DARTS, which again demonstrates the main reason for the poor generalization in DARTS can be the failure of the magnitude-based method.

\begin{table*}[h]
\caption{Comparison of different NAS methods on NAS-Bench-201. Validation and test accuracy with mean and deviation are reported. Our method is repeated 5 times. ``Optimal'' indicates the accuracy of the best architecture in the NAS-Bench-201 search space.}
\label{tab1}
\resizebox{0.98\linewidth}{!}{
\begin{tabular}{lccccccc}
\hline
\multirow{2}{*}{\textbf{Method}}          & \multirow{2}{*}{\textbf{\begin{tabular}[c]{@{}c@{}}Cost\\ (hours)\end{tabular}}} & \multicolumn{2}{c}{\textbf{CIFAR-10}} & \multicolumn{2}{c}{\textbf{CIFAR-100}} & \multicolumn{2}{c}{\textbf{ImageNet-16-120}} \\ \cline{3-8} &  & \textbf{Validation}  & \textbf{Test}  & \textbf{Validation}  & \textbf{Test}   & \textbf{Validation}     & \textbf{Test}   \\ \hline

ResNet~\cite{he2016deep} & -   & 90.83  & 93.97          & 70.42    & 70.86           & 44.53       & 43.63  \\ \hline
DARTS (1st) ~\cite{liu2018darts} & 3.2      & 39.77 $\pm$ 0.00    & 54.30 $\pm$ 0.00         & 15.03 $\pm$ 0.00    & 15.61 $\pm$ 0.00           & 16.43 $\pm$ 0.00     & 16.32 $\pm$ 0.00  \\
DARTS (2nd) ~\cite{liu2018darts} & 9.9      & 39.77 $\pm$ 0.00    & 54.30 $\pm$ 0.00          & 15.03 $\pm$ 0.00    & 15.61 $\pm$ 0.00           & 16.43 $\pm$ 0.00       & 16.32 $\pm$ 0.00  \\
DSNAS\cite{hu2020dsnas} & - & 89.66 $\pm$ 0.29 & 93.08 $\pm$ 0.13 & 30.87 $\pm$ 16.40 & 31.01 $\pm$ 16.38 & 40.61 $\pm$ 0.09 & 41.07 $\pm$ 0.09 \\
SNAS\cite{xie2019snas} & - & 90.10 $\pm$ 1.04 & 92.77 $\pm$ 0.83 & 69.69 $\pm$ 2.39 & 69.34 $\pm$ 1.98 & 42.84 $\pm$ 1.79 & 43.16 $\pm$ 2.64 \\
ENAS ~\cite{pham2018efficient}   & 3.9      & 37.51 $\pm$ 3.19    & 53.89 $\pm$ 0.58          & 13.37 $\pm$ 2.35    & 13.96 $\pm$ 2.33           & 15.06 $\pm$ 1.95       & 14.84 $\pm$ 2.10  \\
SETN ~\cite{dong2019one}         & 9.5      & 84.04 $\pm$ 0.28    & 87.64 $\pm$ 0.00          & 58.86 $\pm$ 0.06    & 59.05 $\pm$ 0.24           & 33.06 $\pm$ 0.02       & 32.52 $\pm$ 0.21  \\
RFGIAug~\cite{xie2023architecture}        & -      & 91.43    & 94.25          & -    &           & -      & -  \\
GDAS ~\cite{dong2019searching}   & 8.8      & 89.89 $\pm$ 0.08    & 93.61 $\pm$ 0.09          & 71.34 $\pm$ 0.04    & 70.70 $\pm$ 0.30           & 41.59 $\pm$ 1.33       & 41.71 $\pm$ 0.98  \\
DARTS- ~\cite{chu2021darts-}     & 3.2      & 91.03 $\pm$ 0.44    & 93.80 $\pm$ 0.40          & 71.36 $\pm$ 1.51    & 71.53 $\pm$ 1.51           & 44.87 $\pm$ 1.46       & 45.12 $\pm$ 0.82  \\ 

NASWOT $\pm$ N=1000 ~\cite{mellor2021neural}    & 0.09          & 89.69 $\pm$ 0.73     & 92.96 $\pm$ 0.81    & 69.86 $\pm$ 1.21          & 69.98 $\pm$ 1.22   & 43.95 $\pm$ 2.05           & 44.44 $\pm$ 2.10   \\
DARTS+PT ~\cite{wang2021rethinking}     & -      & -    & 88.11          & -    & -           & -    & -  \\
iDARTS \cite{zhang2021idarts}    & -     & 89.96 $\pm$ 0.60     & 93.58 $\pm$ 0.32    &  70.57 $\pm$ 0.24         & 70.83 $\pm$ 0.48   & 40.38 $\pm$ 0.59           & 40.89 $\pm$ 0.68   \\
TNAS \cite{qian2022meets}    & 3.6      & -     & 94.35 $\pm$ 0.03    & -          & 73.02 $\pm$ 0.34   & -           & 46.31 $\pm$ 0.24   \\  \hline
\textbf{\nameshortv{}}       & 3.2       & 90.74 $\pm$ 0.66     & 94.02 $\pm$ 0.25 & 72.93 $\pm$ 1.11   & 73.09 $\pm$ 0.77  & 43.02 $\pm$ 2.74      & 44.38 $\pm$ 2.40  \\
\textbf{\nameshort{}}        & 1.2         & \textbf{91.58 $\pm$ 0.00}       & \textbf{94.36 $\pm$ 0.00} & \textbf{73.49 $\pm$ 0.00}       & \textbf{73.51 $\pm$ 0.00}  & \textbf{46.37 $\pm$ 0.00}          & \textbf{46.34 $\pm$ 0.00}     \\
\hline
Random& -        & 83.20 $\pm$ 13.28    & 86.61 $\pm$ 13.46          & 60.70 $\pm$ 12.55    & 60.83 $\pm$ 12.58           & 33.34 $\pm$ 9.39       & 31.13 $\pm$ 9.66  \\
\textbf{Optimal}  & -        & \textbf{91.61}       & \textbf{94.37} & \textbf{73.49}       & \textbf{73.51}  & \textbf{46.77}          & \textbf{47.31}     \\ \hline
\end{tabular}
}
\end{table*}
 \vspace{-5pt}
\section{Experiments}
\label{sec:exp}

\subsection{Experiments on NAS-Bench-201 search space}
\label{exp1}

We first evaluate our methods on NAS-Bench-201~\cite{dong2020bench}. The test performances for all candidate architectures on CIFAR-10/100~\cite{krizhevsky2009learning}, ImageNet-16-120~\cite{chrabaszcz2017downsampled}) were reported. NAS-Bench-201 provides a standard cell-based search space, where each cell contains 6 edges with 5 candidate operations (\textit{none, average pooling, 3x3 and 1x1 convolution, skip connection}). There are $5^6=15625$ candidates in total. The architecture contains three stages connected by a basic residual block~\cite{he2016deep} with a stride of 2 between them, and each cell was stacked five times to build the stage.

\subsubsection{Implementation Details.}

According to~\cite{dong2020bench}, we first search for the optimal cell architecture in the search stage, and then report the ground truth performance. All searching is conducted on CIFAR-10 and then the obtained final architecture is tested on three datasets in NAS-Bench-201. We follow the implementation details as~\cite{ye2022b}.

\subsubsection{Comparison results.}
Table~\ref{tab1} shows the experimental results on NAS-Bench-201. The experiments are conducted 5 times with different random seeds to obtain the final statistical results. Remarkably, our methods can outperform other related NAS methods and can find almost the best architecture among three different datasets. For robustness, \nameshort{} can find the same optimal solution at each searching time, where the performance of the obtained architectures is close to the optimal performance of NAS-Bench-201. Moreover, as shown in Figure~\ref{fig_ab}, the search process of the proposed method can quickly reach its optimal point, e.g., the resulting architecture will not change after around 25 epochs on CIFAR-10. 

We also find that the \nameshort{} can achieve higher accuracy on three datasets with lower standard deviation compared to \nameshortv{}. This demonstrates introducing the magnitude of architecture parameters in the architecture selection criterion actually benefits the selection method. 

\begin{table*}[htbp]
\caption{Comparison of different NAS methods on CIFAR-10/100 in DARTS search space. Test accuracy with mean and deviation are reported. Our method is repeated 5 times.}
\label{tab2}
\centering
\resizebox{1.0\linewidth}{!}{
\begin{tabular}{p{2.4cm}p{2cm}<{\centering}p{2cm}<{\centering}p{1.4cm}<{\centering}p{1.1cm}<{\centering}p{1.1cm}<{\centering}p{1.7cm}<{\centering}p{1.5cm}<{\centering}}
\toprule
\multirow{2}{*}{\textbf{Method}} & \multicolumn{2}{c}{\textbf{Test Acc. (\%)}} & \multirow{2}{*}{\textbf{\begin{tabular}[c]{@{}c@{}}Search Costs\\ (GPU days)\end{tabular}}}  & \multirow{2}{*}{\textbf{\begin{tabular}[c]{@{}c@{}}FLOPs\\ (M)\end{tabular}}} & \multirow{2}{*}{\textbf{\begin{tabular}[c]{@{}c@{}}Params\\ (M)\end{tabular}}} & \textbf{Search} &  \textbf{Selection}  \\ \cline{2-3}
& \textbf{CIFAR-10} & \textbf{CIFAR-100} &  &  &  & \textbf{Method} & \textbf{Method} \\ \midrule


NASNet-A~\cite{zoph2017neural}& 97.35    & 82.19   & 1800   & -& 3.3&RL  & -\\ 

ENAS~\cite{pham2018efficient}       & 96.46    & 80.57     &  0.45     & -& 4.6       &RL  & -\\ 

MetaQNN~\cite{baker2016designing}  & 93.08 & 72.86 & 80 & -& 11.2 & RL& -\\ 
AmoebalNet-B~\cite{real2019regularized}        & 97.45 $\pm$ 0.05          & -   & 3150 & - & 2.8 & evolution & - \\ 
Hierarchical NAS~\cite{liu2017hierarchical}    & 96.25 $\pm$ 0.12          & -   & 300 & - & 15.7 & evolution & - \\

Hierarchical NAS~\cite{liu2017hierarchical}    & 96.25 $\pm$ 0.12          & -   & 300 & - & 15.7 & evolution & -  \\
ModuleNet~\cite{chen2021modulenet} & 97.23 & 82.01 & 2.0 & - & - & evolution & -  \\

\midrule
DARTS (1st)~\cite{liu2018darts} & 97.00 $\pm$ 0.14  & 82.24  & 0.4  & 501 & 3.4 &gradient &magnitude\\
DARTS (2nd)~\cite{liu2018darts} & 97.34 $\pm$ 0.11  & 82.24  & 1.0  & 528 & 3.4 &gradient&magnitude\\
DARTS-~\cite{chu2021darts-}     & 97.41 $\pm$ 0.08  & -& 0.4 & -& 3.5 & gradient &magnitude \\
SDARTS-ADV~\cite{chen2020stabilizing} & 97.39 $\pm$0.02  & -& 1.3 & -& 3.3 & gradient&magnitude \\

SNAS~\cite{xie2019snas} & 97.15 $\pm$ 0.02  & 79.91 & 1.5 & 422 & 2.9 & gradient&magnitude \\
GDAS~\cite{dong2019searching} & 97.07  & 81.62  & 0.21 & 519& 3.4 & gradient&magnitude \\
GDAS-NSAS-C~\cite{zhang2020one} &97.30 $\pm$ 0.07   & 83.30 $\pm$ 0.08  & 0.4 & 520 & 3.3 & gradient&magnitude \\


P-DARTS~\cite{chen2019pdarts}    & 97.50  & 84.08  & 0.3& 532& 3.4 & gradient &magnitude \\
PC-DARTS~\cite{xu2021partially}   & 97.43 $\pm$ 0.07  & 82.89& 0.1 & 557 & 3.6 & gradient &magnitude\\
DropNAS~\cite{hong2020dropnas}    & 97.42 $\pm$ 0.14  & 83.05  $\pm$ 0.14 & 0.7& - & 4.1 & gradient&magnitude \\
DARTS+~\cite{liang2019darts+} & 97.50 $\pm$ 0.11 & 83.72 & 0.4 & - & 3.7 & gradient &magnitude\\ 
$\beta$-DARTS~\cite{ye2022b}    & 97.49 $\pm$ 0.07  & 83.48  $\pm$ 0.03 & 0.4& - & 3.8 & gradient &magnitude\\ 
SWD-NAS~\cite{xue2024self} & 97.49 & 82.92 & 0.13 & - & 3.2 & gradient & magnitude \\
DARTS+PT~\cite{wang2021rethinking}     & 97.39 $\pm $0.08    & -   & 0.8  & -& 3.0& gradient &val. acc\\
Shapley-NAS~\cite{xiao2022shapley}     & 97.53 $\pm$ 0.04    & -   & 0.3  & -  & 3.4 & gradient &val. acc\\
DrNAS~\cite{chen2020drnas}     & 97.54 $\pm$ 0.03    & -   & 0.4  & -  & 4.0  & gradient & distribution\\

FreeDARTS~\cite{zhang2021differentiable}     & 97.22 $\pm$ 0.06    & 81.97   & -  & 634  & 3.6  & gradient & $\Delta$loss\\
\midrule
\textbf{\nameshortv{}}& 97.42 $\pm$ 0.08  & 83.47 $\pm$ 0.15 & 0.4 & 577 & 3.8 & gradient & $\Delta$loss\\
\textbf{\nameshort{}}& \textbf{97.58} $\pm$ 0.06 & \textbf{84.22} $\pm$ 0.30 & 0.4 & 545& 3.5 & gradient & $\Delta$loss\\
\textbf{OStr-PC-DARTS} & 97.38 $\pm$ 0.07  & 83.21 $\pm$ 0.10 & 0.1 & 562 & 3.7 & gradient & $\Delta$loss\\
\textbf{\nameshort{} (Aug)}& \textbf{98.02} & \textbf{85.30} & 0.4& 545 & 3.5 & gradient & $\Delta$loss\\ \bottomrule
\end{tabular}
}
\vspace{-5pt}
\end{table*}

\vspace{-10pt}
\subsection{Performance on DARTS search space}
\label{exp2}
We adopt three datasets CIFAR-10, CIFAR-100, and ImageNet~\cite{deng2009imagenet} for evaluations in DARTS search space~\cite{liu2018darts}. Similar to \cite{liu2018darts} and \cite{xu2021partially}, the mobile setting is considered for ImageNet, under which the input images are cropped to 224$\times$224 and the total number of FLOPs is limited to 600M during inference. The cell in the DARTS search space is consist of 7 nodes and 14 edges. The whole network contains two different cells: the normal and the reduction one. There are 8 different candidate operations on each edge: \textit{zero, max pooling, average pooling, skip connection, 3x3 and 5x5 separable convolution, 3x3 and 5x5 dilation convolution}. The total number of architectures in DARTS space is $8^{14}\approx 4.4\times 10^{11}$. 

\begin{figure*}[t]
	\centering
	\subfigure[Normal cell found on CIFAR-10 by \nameshort{}.]{
	\includegraphics[width=0.48\textwidth]{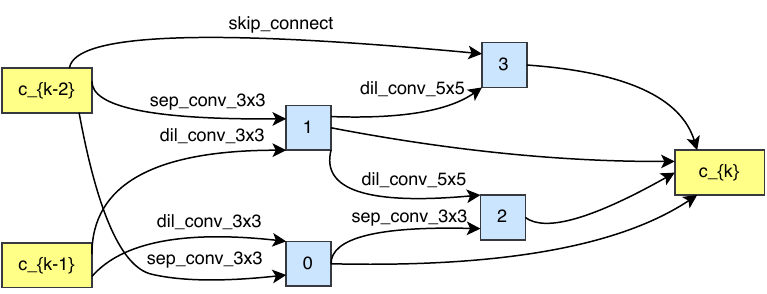}
	}
	\subfigure[Reduction cell found on CIFAR-10 by \nameshort{}.]{
	\includegraphics[width=0.48\textwidth]{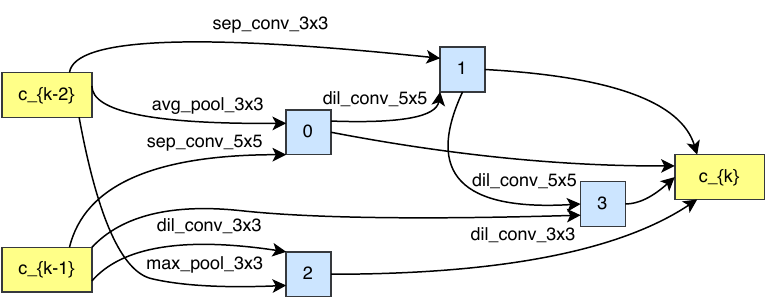}
	}
	\caption{Cells found on CIFAR-10. The normal and reduction cells found by DARTS are dominated with parameter-free operations.}
	\label{fig_vis_darts}
\vspace{-5pt}
\end{figure*}

\begin{table}[htp]
\vspace{-10pt}
\caption{Comparison on ImageNet. All the architectures are searched on CIFAR-10 and evaluated on ImageNet.}
\label{tab3}
\centering
\resizebox{1.0\linewidth}{!}{
\begin{tabular}{p{2.5cm}p{2cm}<{\centering}p{1.2cm}<{\centering}p{1.2cm}<{\centering}}
\hline
\multirow{2}{*}{\textbf{Method}} & \multirow{2}{*}{\textbf{\begin{tabular}[l]{@{}c@{}}Test Acc. (\%) \\ Top-1 / 5\end{tabular}}}  & \multirow{2}{*}{\textbf{\begin{tabular}[c]{@{}c@{}}FLOPs\\ (M)\end{tabular}}} & \multirow{2}{*}{\textbf{\begin{tabular}[c]{@{}c@{}}Params\\ (M)\end{tabular}}} \\ 
&&&  \\ \hline
ShuffleNetV2~\cite{ma2018shufflenet}& 73.7 / - & 524 & 5.0  \\
CondenseNet~\cite{huang2018condensenet}& 73.8 / 91.7 & 529 & 4.8 \\
CondenseNetV2~\cite{yang2021condensenet}& 75.9 / 92.7 & 309 & 6.1 \\ \hline
NASNet-A~\cite{zoph2018learning} & 74.0 / 91.6 & 564 & 5.2 \\
AmoebalNet-A~\cite{real2019regularized}  & 74.5 / 92.0 & 555  & 5.1 \\ 
AmoebalNet-C~\cite{real2019regularized}  & 75.7 / 92.4 & 570 & 6.4 \\ 
RFGIAug~\cite{xie2023architecture}  & 73.4 / - & - & 4.8 \\ 
\hline
DARTS (2nd)~\cite{liu2018darts}   & 73.3 / 91.3 & 574  & 4.9  \\
SNAS~\cite{xie2019snas}  & 72.7 / 90.8 & 522 & 4.3 \\
GDAS~\cite{dong2019searching}  & 72.5 / 91.5  & 497 & 4.4 \\
GDAS-NSAS-C~\cite{zhang2020one}  & 74.1 / -  & 565 & 5.2 \\
PC-DARTS~\cite{xu2021partially} & 74.9 / 92.2  & 586 & 5.3  \\
P-DARTS~\cite{chen2019pdarts} & 75.6 / 92.6 & 557 & 4.9 \\
FairDARTS-B~\cite{chu2020fair}  & 75.1 / 92.5  & 541 &- \\
DARTS+PT~\cite{wang2021rethinking} & 74.5 / 92.0  & -    & -  \\
Shapley-NAS~\cite{xiao2022shapley} & 75.7 / - & 566 & 5.1 \\
DrNAS~\cite{chen2020drnas} & 75.8 / 92.7 & - & 5.2 \\
$\beta$-DARTS~\cite{ye2022b} & 75.8 / 92.9 & 597 & 5.4 
\\\hline
\textbf{\nameshortv{}}  &  75.7 / 92.6 & 598 & 5.4 \\
\textbf{\nameshort{}}  & 76.2 / 93.0  & 550 & 5.0\\
\textbf{OStr-PC-DARTS} & 75.6 / 92.6 & 596 & 5.4 \\
\textbf{\nameshort{} (Aug)} & \textbf{76.7} / \textbf{93.1} & 550 & 5.0 \\
\hline
\end{tabular}
}\vspace{-10pt}
\end{table}

\subsubsection{Implementation Details} 
We follow the experimental settings in~\cite{liu2018darts}: we first search for the optimal cell architecture in the search stage, and then train the resulted architectures from scratch in the evaluation stage. The search is conducted on CIFAR-10 and then the obtained final architecture is tested on both CIFAR-10, CIFAR-100 and ImageNet. The searching and evaluation settings are the same as ~\cite{xu2021partially} and~\cite{liu2018darts}.

To further explore the potential of the obtained architectures, we also train the selected architecture with the augmentation training setting, which applies RandAugmentation~\cite{cubuk2020randaugment} with the double epochs (denoted by (aug) in the tables).

\subsubsection{Results on CIFAR-10 and CIFAR-100}
The results are shown in Table~\ref{tab2}. The experiments are conducted 5 times with different random seeds to obtain the final statistical results for our methods. Figs.~\ref{fig_vis_darts} illustrates the searched normal and reduction cells on CIFAR-10 for \nameshort{}.

From the results, we see that the searched architecture of \nameshort{} can achieve the average accuracy of 97.58\% and 84.22\% on CIFAR-10 and CIFAR-100, respectively, which outperform most of the other methods in Table~\ref{tab2}. Both \nameshort{} and \nameshortv{} outperform the baseline DARTS~\cite{liu2018darts} model with a noticeable gap in terms of accuracy and search stability, which demonstrates the effectiveness of our methods. From the visualization results, we infer that the domination of the parameter-free operations in the searched cell can be the main reason for the performance drop in DARTS. Moreover, the promising performance of \nameshort{} and \nameshortv{} indicates that the improvements for supernet optimization and the subsequent architecture selection criterion are both crucial for designing a high-performance DARTS-based method. 

$\beta$-DARTS~\cite{ye2022b} and PC-DARTS~\cite{xu2021partially} also achieve promising results. We see that both of these methods are superior to \nameshortv{} on CIFAR-10, while having lower accuracy on CIFAR-100 compared with \nameshortv{}. However, \nameshort{} outperforms all the evaluated methods in Table~\ref{tab2} on CIFAR-10 and CIFAR-100. Moreover, after being equipped with RandAugmentation, the performance of \nameshort{} can be further improved to 98.02$\%$ and 85.30$\%$ on CIFAR-10 and CIFAR-100, respectively. 

Our method can also be combined with other searching methods (Ostr-PC-DARTS), such as PC-DARTS~\cite{xu2021partially}. As shown in Table~\ref{tab2}, OStr-PC-DARTS achieves 97.38\% accuracy on CIFAR-10 with the same search costs as PC-DARTS, which is slightly lower than that of PC-DARTS (98.43\%). However, Ostr-PC-DARTS has a stronger generalization ability and outperforms PC-DARTS on CIFAR-100. We also see that OStr-PC-DARTS is significantly superior to PC-DARTS when evaluated on ImageNet.

\begin{table}[htp]
\caption{Comparison with different NAS methods on ImageNet (directly searched on ImageNet).}
\label{tab_im}
\centering
\resizebox{1.0\linewidth}{!}{
\begin{tabular}{p{3.0cm}p{0.9cm}<{\centering}p{0.9cm}<{\centering}p{0.9cm}<{\centering}p{0.9cm}<{\centering}}
\hline
\multirow{2}{*}{\textbf{Method}} & \multirow{2}{*}{\textbf{\begin{tabular}[l]{@{}c@{}}Test Acc.\\Top-1\end{tabular}}} & \multirow{2}{*}{\textbf{\begin{tabular}[c]{@{}c@{}}Search\\ Costs\end{tabular}}} & \multirow{2}{*}{\textbf{\begin{tabular}[c]{@{}c@{}}FLOPs\\ (M)\end{tabular}}} & \multirow{2}{*}{\textbf{\begin{tabular}[c]{@{}c@{}}Params\\ (M)\end{tabular}}} \\ 
&&&  \\ \hline
PC-DARTS~\cite{xu2021partially} & 75.8 & 3.8 & 597 &5.3  \\
EPC-DARTS~\cite{xu2021partially} & 75.7 & 2.8 & 583 & 5.1  \\
FairDARTS~\cite{chu2020fair} & 75.6 & 3.0& 440 & 4.3  \\
RLNAS~\cite{zhang2021neural} & 75.9 & - & 597 & 5.5 \\
DARTS+~\cite{liang2019darts+} & 76.1 & 7.6 & 582 & 5.1 \\ 
 Shapley-NAS~\cite{xiao2022shapley} &  76.1 &  4.2 &  582 &  5.4 \\
 DrNAS~\cite{chen2020drnas} &  76.2 &  4.6 &  - & 5.7 \\
\hline
\textbf{\nameshort{}}  & \textbf{76.3} & 7.6 & 596 & 5.3 \\
\textbf{OStr-PC-DARTS} & 76.2 & 3.8 & 583 & 5.2 \\
\textbf{\nameshort{} (Aug)}  & \textbf{77.1} & 7.6 & 596 & 5.3 \\
\textbf{OStr-PC-DARTS (Aug)} & 76.7 & 3.8 & 583 & 5.2 \\
\hline
\end{tabular}
}
\end{table}

\subsubsection{Results on ImageNet}
\label{sec:imgnet}
We further verify the transferability of the proposed methods by testing the performance of the architecture searched on CIFAR-10. The results are shown in Table~\ref{tab3}, from which we see that both \nameshort{} and \nameshortv{} outperform the original DARTS~\cite{liu2018darts}. The \nameshort{} and \nameshortv{} can achieve 76.2\% and 75.7\% test accuracy on ImageNet, ranking top among the popular NAS methods. Furthermore, we see that the searched architecture of OStr-PC-DARTS is significantly superior to that of PC-DARTS when evaluated on ImageNet. And the resulting architecture of \nameshort{} can achieve 76.7\% test accuracy on ImageNet under the augmentation training setting.

\subsubsection{Direct search on ImageNet}
We further follow the searching settings in~\cite{xu2021partially} and conduct the direct search on ImageNet. Table~\ref{tab_im} shows the evaluated results of different NAS methods directly searching on the ImageNet dataset. We see that the proposed \nameshort{} achieves a top-1 accuracy of 76.3\%, which is superior to other tested baseline models in Table~\ref{tab_im}. Our method can also combine with PC-DARTS to enjoy its fast searching speed. The OStr-PC-DARTS outperforms the original PC-DARTS (76.2\% v.s. 75.8\%) with the same search costs. The test accuracy of \nameshort{} and OStr-PC-DARTS can be further improved to 77.1\% and 76.7\% with the augmented training settings. 


\begin{table}[]
	\caption{Ablation study for different selection criteria.}
	\label{tab_ab}
	\centering
	\resizebox{0.98\linewidth}{!}{
	\begin{tabular}{p{1.8cm}p{1.8cm}<{\centering} p{1.8cm}<{\centering}p{1.8cm}<{\centering}}
	\hline
	\textbf{Method} & Criterion & \textbf{CIFAR-10} & \textbf{CIFAR-100} \\ \hline
	DARTS  &  $\bb$    & 54.30 $\pm$ 0.00     & 15.61 $\pm$ 0.00\\
	Naive  &  Pruning  & 93.86 $\pm$ 0.29     & 71.88 $\pm$ 0.37\\
	\nameshortv{}  &  $s^*$ & 94.02  $\pm$ 0.25   & 73.09 $\pm$ 0.77\\
	\nameshort{} &  $s$ & \textbf{94.36 $\pm$ 0.00}    & \textbf{73.51 $\pm$ 0.00} \\
	\hline
	\end{tabular}
	}
\end{table}

\begin{table*}
\caption{Evaluations on the Stability of DARTS, \nameshort{} and \nameshortv{}. std means standard deviation.}
\label{tab_stab}
\centering
\resizebox{0.95\textwidth}{!}{
\begin{tabular}{p{2.8cm}p{1cm}<{\centering}p{1cm}<{\centering}p{1cm}<{\centering}p{1cm}<{\centering}p{1cm}<{\centering}p{1cm}<{\centering}p{1cm}<{\centering}p{1cm}<{\centering}p{1cm}<{\centering}p{1cm}<{\centering}}
\hline
\multirow{3}{*}{\textbf{Methods}} & \multicolumn{10}{c}{\textbf{Test Acc. (\%)}}           \\ \cline{2-11}
     & \multicolumn{6}{c}{\textbf{Search Runs}} & \multicolumn{4}{c}{\textbf{Number of Epochs}} \\ \cmidrule(r){2-7} \cmidrule(r){8-11} 
 & 1      & 2      & 3      & 4     & 5   & std  & 25        & 50        & 100   & std    \\ \cmidrule(r){2-7} \cmidrule(r){8-11} 
DARTS(1st order)      & 97.11  & 96.85  & 96.93  & 97.01 & 96.73 & 0.15    & 97.26     & 96.96     & 96.52   & 0.37  \\
DARTS(2nd order)      & 97.12  & 97.04  & 97.30  & 97.24 & 97.08 & 0.11     & 97.32     & 97.15     & 96.94 & 0.19   \\
\nameshort{}  & 97.43  & 97.63  & 97.59  & 97.58 & 97.52 & 0.06 & 97.53     & 97.55     & 97.43   &0.06  \\
\nameshortv{} & 97.50  & 97.43  & 97.38  & 97.48 & 97.39 & 0.08     & 97.45     & 97.44     & 97.42 &0.02    \\ \hline
\end{tabular}%
}
\end{table*}

\begin{figure}
	\centering
	\subfigure[CIFAR-10.]{
	\label{fig:a} 
	\includegraphics[width=0.22\textwidth]{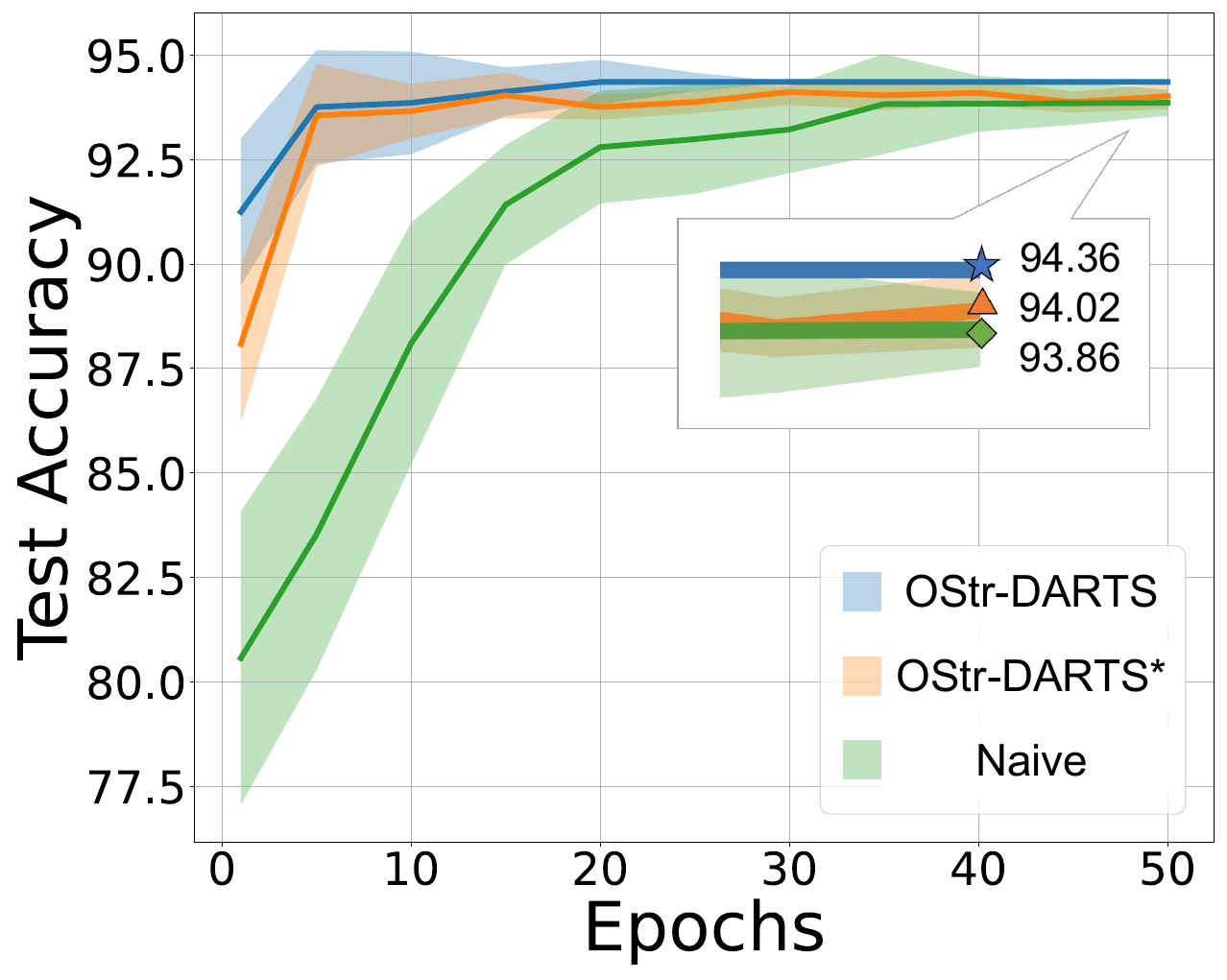}
	}	\subfigure[CIFAR-100.]{
	\label{fig:b} 
	\includegraphics[width=0.22\textwidth]{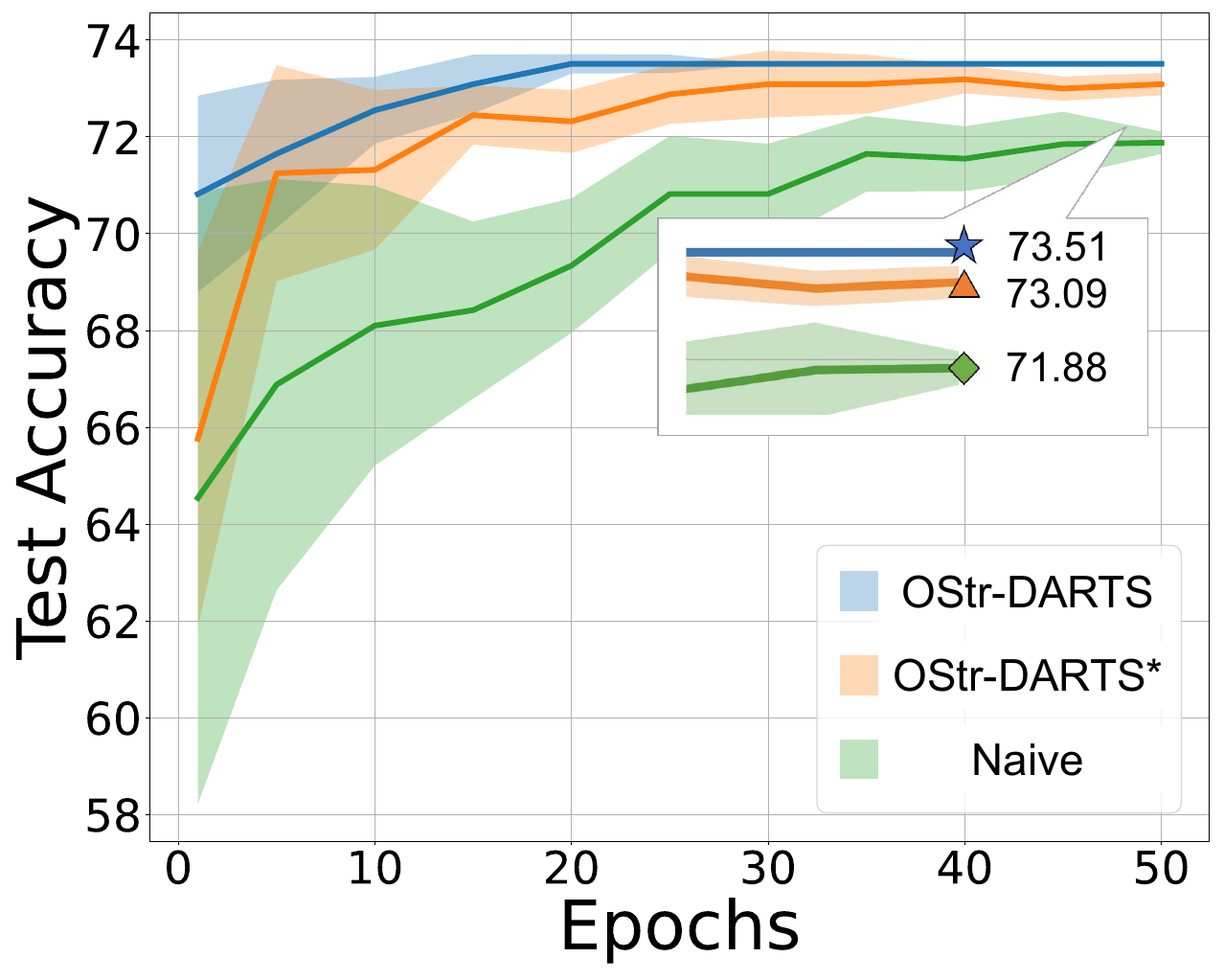}
	}
	\caption{Test accuracy of the selected architectures with three implementations at epochs in NAS-Bench-201 search space.}
	\label{fig_ab}
\end{figure}

\begin{figure}[t]
	\centering
	\subfigure[CIFAR-10.]{
	\includegraphics[width=0.22\textwidth]{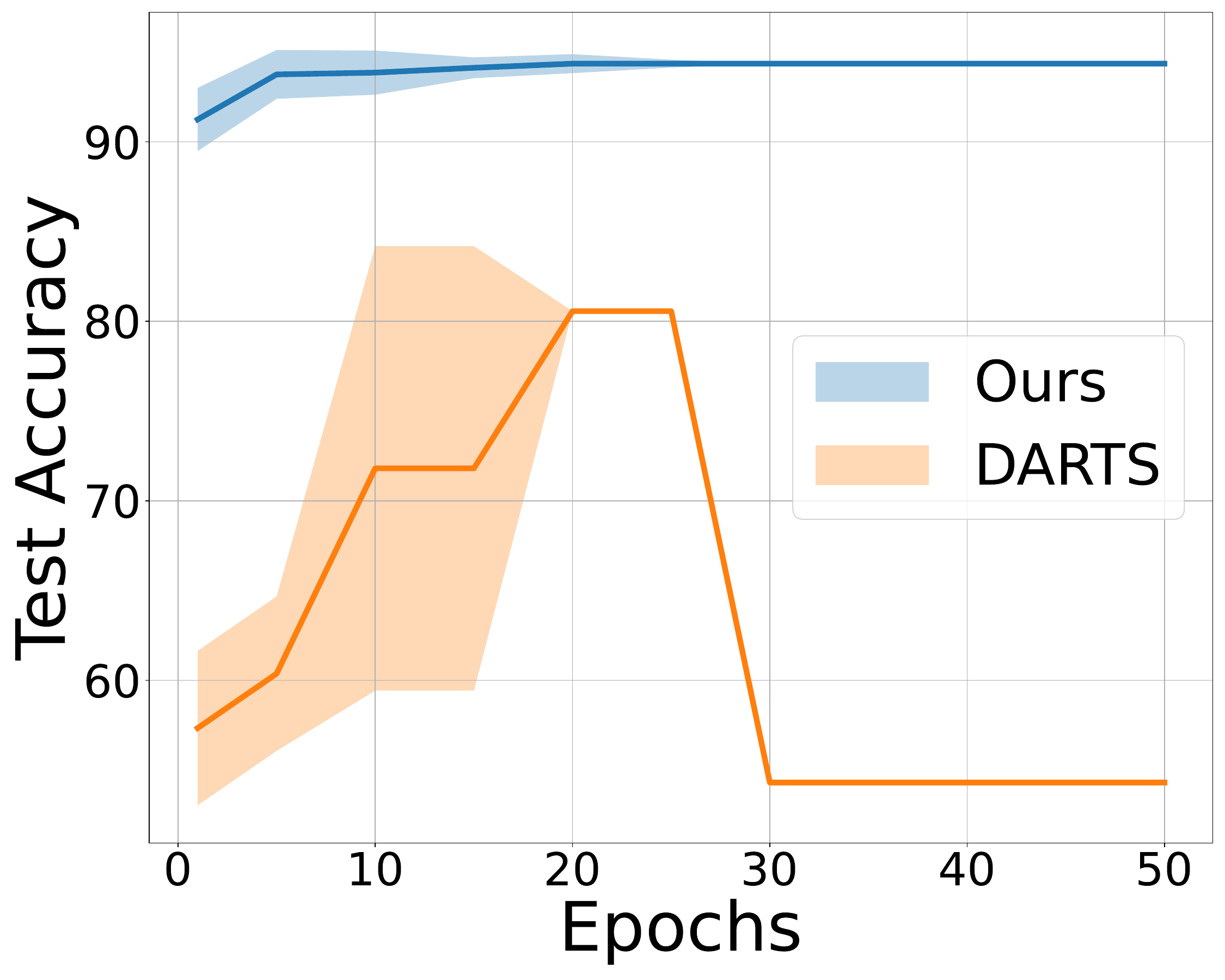}
	}	
	\subfigure[CIFAR-100.]{
	\includegraphics[width=0.22\textwidth]{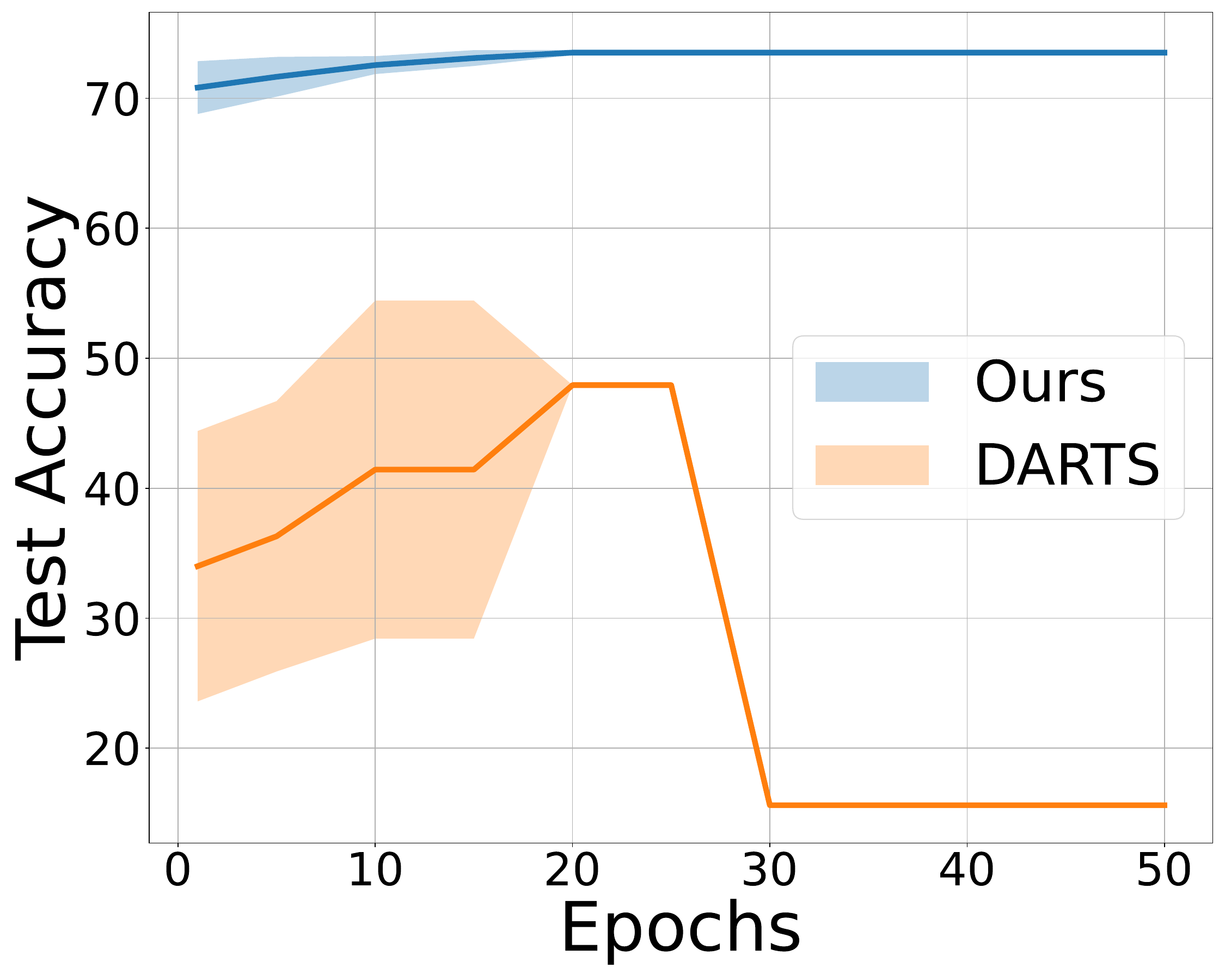}
	}
	\caption{Test accuracy of selected architectures of DARTS and \nameshort{} at different epochs on (a) CIFAR-10 and (b) CIFAR-100.}
	\label{fig_degne}
\end{figure}

\begin{figure}[htp]
	\centering
	\includegraphics[width=0.48\textwidth]{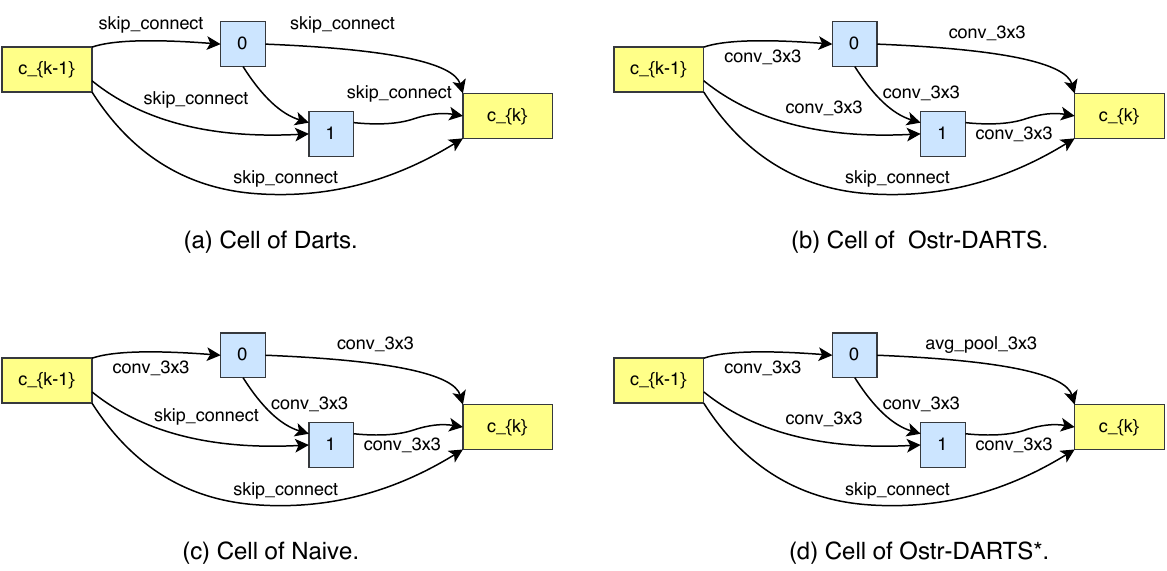}
	\caption{Visualization results of (a) DARTS, (b) \nameshort{}, (c) Naive and (d) \nameshortv{}.}
	\label{fig_201vis}
 \vspace{-10pt}
\end{figure}

\subsection{Ablation Study}
This section provides a comprehensive study of the proposed method. We first show that the degeneration issue can be effectively addressed by using the proposed selection criteria, indicating the importance of the architecture selection method in DARTS. Then we study the optimization stability of DARTS and \nameshort{}. We further test the stand-alone model performance to evaluate the correlation between the selection indicator and the final performance.

\subsubsection{The importance of architecture selection}

From the results in Fig.~\ref{fig_degne}, we see that the degeneration problem can be easily addressed by using operation strength. Moreover, to further confirm our results, Table~\ref{tab_ab} and Fig.~\ref{fig_ab} evaluate the performance of the DARTS methods based on three different select criteria: 1) the operation strength, $s$; 2) the variant of operation strength* stated in Section IV-D, $s^*$; 3) the Naive pruning based method stated in Fig.~\ref{fig_3} (b). From the results, we see that the performance of the selected architectures will not drop with the increase of searching epochs (Fig.~\ref{fig_ab}). Note that DARTS, Naive, \nameshortv{}, \nameshort{}, use the same optimized supernet, we indicate that the magnitude-based architecture selection method can be a critical reason for the instability of classical DARTS, and the degeneration issue in DARTS can be effectively addressed by an appropriate architecture selection criterion without any improvement for supernet optimization.

Moreover, we see that the direct implementation of the pruning methods to estimate the importance of operations can result in both slower coverage speed of the searching process and lower performance of the final architecture. This indicates that developing the estimation method aiming at the selection procedure is crucial for the importance estimation of operation selection. However, due to the better estimation accuracy of \nameshort{}, its performance is superior to \nameshortv{} in the experiments.

\subsubsection{Optimization stability}

Table~\ref{tab_stab} evaluates the top-1 test error rates of the architectures searched by \nameshort{}, \nameshortv{} and DARTS under different runs and numbers of epochs. The architectural parameters of all methods are generated by random seeds that vary for the runs. Overall, DARTS with first- and second-order approximation are less stable. For ours, the standard deviations are 0.06 and 0.08, which are better than DARTS. Methods with 25, 50, and 100 searching epochs are also evaluated. Due to the degeneration problem, top-1 test accuracy rates of DARTS would drop after 25 epochs, since DARTS tends to select the skip-connections with more epochs~\cite{liang2019darts+}. The standard deviation of the accuracy with different epochs can be up to 0.37, which is much higher than those of the proposed \nameshort{} and \nameshortv{}.

From Fig. \ref{fig_degne}, we observe that the performance of the resulted architecture by DARTS starts to degenerate after around 25 epochs of searching. The visualization in Fig. \ref{fig_201vis} (a) indicates that such a performance collapse is mainly due to the domination of skip connection. While the proposed \nameshort{} can effectively address the degeneration problem and has a stable searching process. The searching process tends to be stable after around 25 epochs.

\subsubsection{Stand-Alone Model Performance}

\begin{figure}
	\centering
	\subfigure[Stand-Alone Model Performance on Edge-0.]{
	\includegraphics[width=0.48\textwidth]{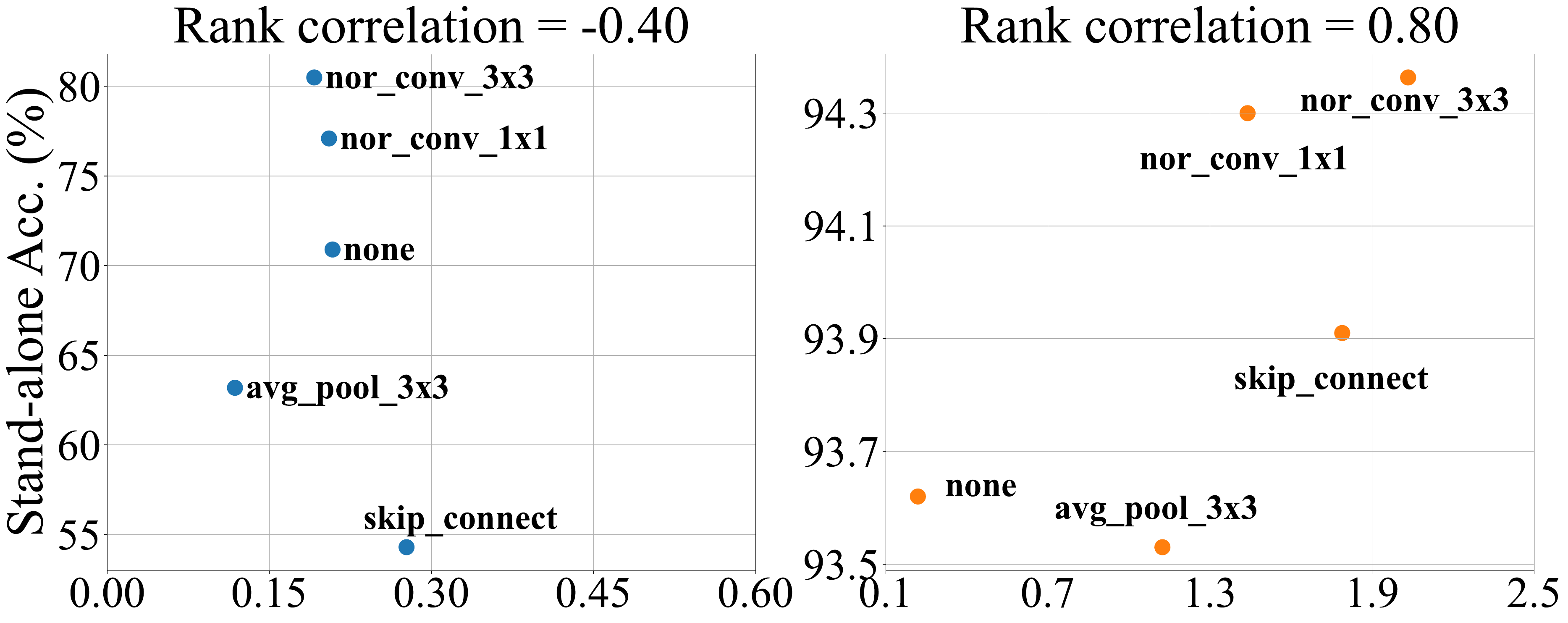}
        \label{fig_saa}
	}
	\subfigure[Stand-Alone Model Performance on Edge-3.]{
	\includegraphics[width=0.48\textwidth]{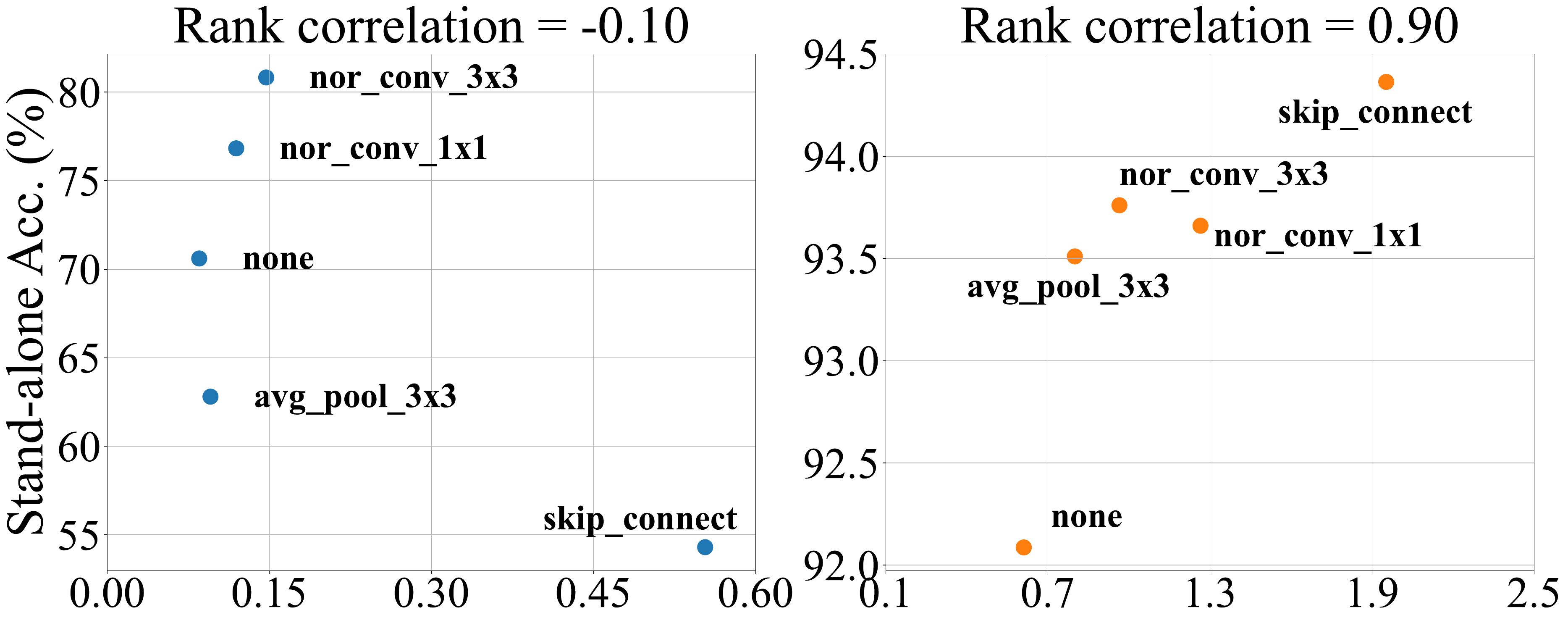}
	}
	\caption{Correlation between the accuracy of standalone model and their corresponding indicators ($\alpha$ for DARTS, operation strength for \nameshort{}) on NAS-Bench-201 search space. }
	\label{fig_sa}
 \vspace{-10pt}
\end{figure}

To evaluate the optimality of the selected operations, we use Spearman's rank correlation coefficient to evaluate the relationships between the applied indicators and their corresponding stand-alone model performance. After the searching and selection phrase of the supernet, we obtain the final subnet (final target network) of one chosen operation at each edge, and the stand-alone performance is calculated as: 1) varying operations at one certain edge (Edge-0 in Figure \ref{fig_saa}) in the candidate set while keeping operation choices at other edges unchanged; 2) retraining the ablated subnet and obtain the corresponding stand-alone performance standing for the ground-truth evaluation of operation selection. Fig.~\ref{fig_sa} (a) and (b) show the correlation between the applied indicators and the corresponding stand-alone model accuracy on Edge-0 and Edge-3 (The cell architecture is shown in Fig.~\ref{fig_cell}.) in the NAS-Bench-201 search space. The architecture parameters $\alpha$ and the accuracy of the corresponding final networks have with weak correlation, indicating the serious degeneration problem of DARTS and the limitation of the magnitude-based architecture selection method. However, the correlations between the accuracy and operation strength are much higher in our methods. The operations with the largest operation strength correspond to the stand-alone models with the highest accuracy, indicating that operation strength can correctly reflect the contribution of an operation to the supernet compared with the magnitude of architecture parameters.

\begin{figure}
	\centering
	\includegraphics[width=0.48\textwidth]{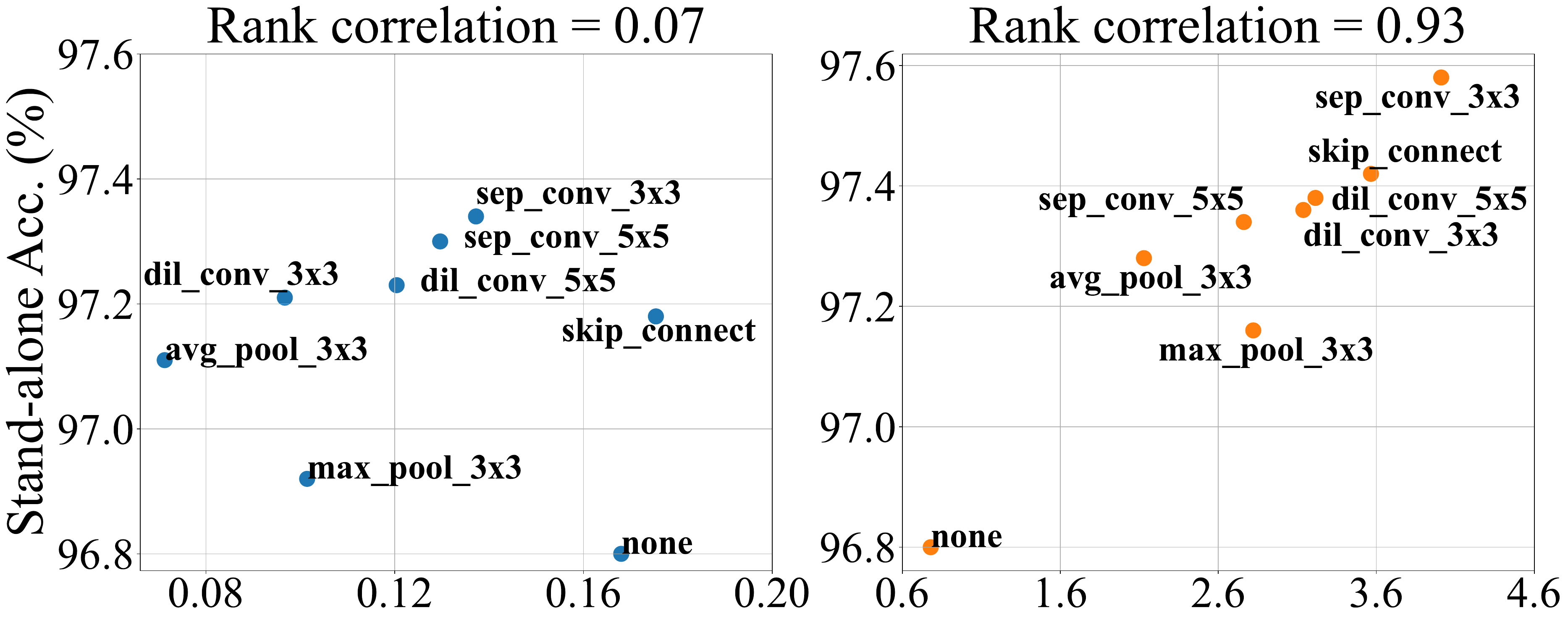}
	\caption{Correlation between the accuracy of standalone model and their corresponding indicators ($\alpha$ for DARTS, operation strength for \nameshort{}) on DARTS search space. }
	\label{fig_sa_darts}
 \vspace{-10pt}
\end{figure}

\begin{table*}[htp]
\caption{Object detection on MS COCO for RetinaNet implemented with the backbone obtained by different DARTS-based methods.}
\label{tab_ob}
\centering
\resizebox{0.95\linewidth}{!}{
\begin{tabular}{p{3cm}p{1.6cm}<{\centering}p{1.6cm}<{\centering}p{1.1cm}<{\centering}p{1.1cm}<{\centering}p{1.1cm}<{\centering}p{1.1cm}<{\centering}p{1.1cm}<{\centering}p{1.1cm}<{\centering}}
\hline
\textbf{Method} & \textbf{Params (M)} & \textbf{FLOPs (M)} & AP & AP$_{50}$ & AP$_{75}$ & AP$_{S}$ & AP$_{M}$ & AP$_{L}$ \\ \hline
Random search & 4.7 & 519 & 31.7 & 50.4 & 33.4 & 16.3 & 35.2 & 42.9 \\ \hline
DARTS (2nd order)~\cite{liu2018darts}  & 4.9 & 574 & 31.4 & 49.0 & 33.2 & 16.3 & 33.7 & 43.4 \\
P-DARTS~\cite{chen2019pdarts}  & 4.9 & 557 & 32.9 & 52.1 & 34.7 & 17.2 & 36.2 & 44.8 \\
PC-DARTS~\cite{xu2021partially}  & 5.3 & 586 & 32.9 & 51.8 & 34.8 & 17.5 & 36.3 & 43.5 \\
UnNAS(color)~\cite{liu2020labels}  & 5.3 & 587 & 32.4 & 51.2 & 34.2 & 16.6 & 35.6 & 44.6 \\
UnNAS(jigsaw)~\cite{liu2020labels}  & 5.2 & 560 & 33.0 & 51.9 & 35.3 & 16.4 & 37.2 & 45.4 \\
RLNAS~\cite{zhang2021neural}  & 5.5 & 597 & 32.4 & 50.9 & 34.4 & 16.5 & 35.5 & 44.5 \\ \hline
\textbf{\nameshort{}} & 5.0 & 550 & \textit{34.5} & \textit{53.1} & \textit{36.4} & \textit{17.8} & \textit{37.5} & \textit{47.9} \\ 
\hline
\end{tabular} 
}
\vspace{-10pt}
\end{table*}
The correlation in the DARTS space shows a similar result (Fig.~\ref{fig_sa_darts}). We see that the architecture parameters $\alpha$ and the accuracy of the corresponding final networks have with weak correlation (0.07). However, the rank correlations between the accuracy of the stand-alone model and operation strength are much higher (0.93).

\subsection{Transferring to Object Detection}
The obtained neural architectures are further used as the backbone networks in object detection tasks. We integrate the searched architecture with RetinaNet~\cite{lin2017focal}, a popular single-stage detection framework. The detection models are trained on the train set of MS-COCO~\cite{lin2014microsoft} and tested on the validation set. We follow the standard training setting in~\cite{lin2017focal} and only modify the backbone of RetinaNet. 

Table~\ref{tab_ob} summarizes the average precision (AP) metrics of object detection, including AP averaged over different IoU thresholds (AP) and AP for small, medium and large objects (AP$_S$, AP$_M$ and AP$_L$). The FLOPs in the table mean the backbone FLOPs measured on ImageNet with 224$\times$224. RetinaNet is implemented with the backbone architectures searched by different DARTS-based methods, which are pre-trained on ImageNet. From the results, we see that \nameshort{} achieves an AP of 34.4 percent and outperforms all other methods in Table\ref{tab_ob}. These results suggest that \nameshort{} can maintain the strong generalization ability when transferred to other vision tasks such as object detection. Therefore, we believe that building stronger architecture selection methods for DARTS-based methods would benefit more application scenarios to explore the high-performance neural architectures in an Auto-ML manner.

\section{Limitations and future works}
\label{lim}
While experimental results show the effectiveness of \nameshort{}, our method still has some limitations. 

First, the optimality of the proposed method can only be guaranteed with the assumption that the change of loss can be used to estimate the solution of the discrete NAS selection problem in (\ref{a_eq1}). Although such a solution method is effective in pruning fields~\cite{lee2019signal}, how good it is should be further investigated in DARTS. However, the experiments in Fig.~\ref{fig_sa} and Fig.~\ref{fig_sa_darts} show a high correlation between the accuracy of the final model and the selected operations, indicating that using operation strength can be a possible way for approximating the solution of the NAS selection problem.

Second, the optimality of the proposed method can only be guaranteed with the assumption that the operation strength can be used to estimate the loss changes when a certain operation is selected. As DARTS is notorious for its precipitous validation loss landscape, whether the proposed operation strength can be used to accurately estimate the loss change still needs to be further studied.

Third, our method relies on approximating operation strength from gradient information. Considering the precipitous validation loss landscape, the gradient information might sometimes with high variance. We believe that using the second-order Taylor term might lead to a better loss change estimator for the architecture selection problem. However, we only apply the first-order Taylor expansion for loss change estimation due to the high computational costs of calculating the Hessian Matrix.

As we do not apply any improvements to the supernet optimization of DARTS, we indicate that the poor generalization observed in DARTS can be attributed to the failure of magnitude-based architecture selection rather than the optimization of its supernet entirely. Our research found that besides architecture parameters, the generated features might also be a key factor in identifying the important connections or operations in a neural network. Recent work~\cite{sun2023simple} has also proposed a similar pruning strategy that considers both weights and activation values during pruning Large Languages Models (LLM) and achieved promising performance. Future work mainly includes building the sophisticated importance estimation criteria to serve the NAS community as well as the pruning and large model fields, or even NAS for LLM. We believe that all of them will contribute a lot to building high-performance neural architectures with less complexity.

\section{Conclusion}
\label{sec:conclusion}
In this paper, we propose a novel architecture selection criterion for DARTS-based methods, based on which we further develop an effective NAS approach named \nameshort{}. The new architecture selection method measures the contribution of an operation to the supernet by introducing the effect of this operation on the final loss. We find that the frequent degeneration issue in DARTS can be effectively addressed by simply substituting the original magnitude-based selection method with ours. Moreover, the proposed selection criterion can be combined with various orthogonal improvements for supernet optimization if necessary. Experimental results show that our methods can consistently explore outperformed architectures from supernets compared to other related baseline models on several search spaces and datasets. We hope that our work can bring a new perspective to the NAS community to design high-performance DARTS-based methods by developing sophisticated architecture selection methods from the pre-trained or scratch DARTS supernet.
\vspace{5pt}

\small{
\bibliographystyle{IEEEtran}
\bibliography{reference}

\begin{thebibliography}{10}
\providecommand{\url}[1]{#1}
\csname url@samestyle\endcsname
\providecommand{\newblock}{\relax}
\providecommand{\bibinfo}[2]{#2}
\providecommand{\BIBentrySTDinterwordspacing}{\spaceskip=0pt\relax}
\providecommand{\BIBentryALTinterwordstretchfactor}{4}
\providecommand{\BIBentryALTinterwordspacing}{\spaceskip=\fontdimen2\font plus
\BIBentryALTinterwordstretchfactor\fontdimen3\font minus \fontdimen4\font\relax}
\providecommand{\BIBforeignlanguage}[2]{{%
\expandafter\ifx\csname l@#1\endcsname\relax
\typeout{** WARNING: IEEEtran.bst: No hyphenation pattern has been}%
\typeout{** loaded for the language `#1'. Using the pattern for}%
\typeout{** the default language instead.}%
\else
\language=\csname l@#1\endcsname
\fi
#2}}
\providecommand{\BIBdecl}{\relax}
\BIBdecl

\bibitem{zoph2017neural}
B.~Zoph and Q.~V. Le, ``Neural architecture search with reinforcement learning,'' \emph{Int. Conf. Learn. Represent.}, 2017.

\bibitem{zoph2018learning}
B.~Zoph, V.~Vasudevan, J.~Shlens, and Q.~V. Le, ``Learning transferable architectures for scalable image recognition,'' in \emph{IEEE Conf. Comput. Vis. Pattern Recog.}, 2018, pp. 8697--8710.

\bibitem{baker2016designing}
B.~Baker, O.~Gupta, N.~Naik, and R.~Raskar, ``Designing neural network architectures using reinforcement learning,'' \emph{Int. Conf. Learn. Represent.}, 2017.

\bibitem{real2019regularized}
E.~Real, A.~Aggarwal, Y.~Huang, and Q.~V. Le, ``Regularized evolution for image classifier architecture search,'' in \emph{Proceedings of the AAAI conference on artificial intelligence}, vol.~33, no.~01, 2019, pp. 4780--4789.

\bibitem{liu2017hierarchical}
H.~Liu, K.~Simonyan, O.~Vinyals, C.~Fernando, and K.~Kavukcuoglu, ``Hierarchical representations for efficient architecture search,'' in \emph{Int. Conf. Learn. Represent.}, 2018.

\bibitem{chen2021modulenet}
Y.~Chen, R.~Gao, F.~Liu, and D.~Zhao, ``Modulenet: Knowledge-inherited neural architecture search,'' \emph{{IEEE} Trans. Cybern.}, vol.~52, no.~11, pp. 11\,661--11\,671, 2021.

\bibitem{zhang2020one}
M.~Zhang, H.~Li, S.~Pan, X.~Chang, C.~Zhou, Z.~Ge, and S.~Su, ``One-shot neural architecture search: Maximising diversity to overcome catastrophic forgetting,'' \emph{{IEEE} Trans. Pattern Anal. Mach. Intell.}, vol.~43, no.~9, pp. 2921--2935, 2020.

\bibitem{dong2019one}
X.~Dong and Y.~Yang, ``One-shot neural architecture search via self-evaluated template network,'' in \emph{Int. Conf. Comput. Vis.}, 2019, pp. 3680--3689.

\bibitem{pham2018efficient}
H.~Pham, M.~Guan, B.~Zoph, Q.~Le, and J.~Dean, ``Efficient neural architecture search via parameters sharing,'' in \emph{Int. Conf. Mach. Learn.}, vol.~80, 2018, pp. 4095--4104.

\bibitem{liu2018darts}
H.~Liu, K.~Simonyan, and Y.~Yang, ``Darts: Differentiable architecture search,'' in \emph{Int. Conf. Learn. Represent.}, 2019.

\bibitem{zela2019understanding}
A.~Zela, T.~Elsken, T.~Saikia, Y.~Marrakchi, T.~Brox, and F.~Hutter, ``Understanding and robustifying differentiable architecture search,'' in \emph{Int. Conf. Learn. Represent.}, 2020.

\bibitem{wang2021rethinking}
R.~Wang, M.~Cheng, X.~Chen, X.~Tang, and C.-J. Hsieh, ``Rethinking architecture selection in differentiable nas,'' in \emph{Int. Conf. Learn. Represent.}, 2021.

\bibitem{hong2020dropnas}
W.~Hong, G.~Li, W.~Zhang, R.~Tang, Y.~Wang, Z.~Li, and Y.~Yu, ``Dropnas: Grouped operation dropout for differentiable architecture search.'' in \emph{International Joint Conferences on Artificial Intelligence (IJCAI)}, 2020.

\bibitem{gu2021dots}
Y.-C. Gu, L.-J. Wang, Y.~Liu, Y.~Yang, Y.-H. Wu, S.-P. Lu, and M.-M. Cheng, ``{DOTS}: Decoupling operation and topology in differentiable architecture search,'' in \emph{IEEE Conf. Comput. Vis. Pattern Recog.}, 2021, pp. 12\,306--12\,315.

\bibitem{liang2019darts+}
H.~Liang, S.~Zhang, J.~Sun, X.~He, W.~Huang, K.~Zhuang, and Z.~Li, ``Darts+: Improved differentiable architecture search with early stopping,'' \emph{arXiv preprint arXiv:1909.06035}, 2019.

\bibitem{ye2022b}
P.~Ye, B.~Li, Y.~Li, T.~Chen, J.~Fan, and W.~Ouyang, ``b-darts: Beta-decay regularization for differentiable architecture search,'' in \emph{IEEE Conf. Comput. Vis. Pattern Recog.}, 2022.

\bibitem{chen2020stabilizing}
X.~Chen and C.-J. Hsieh, ``Stabilizing differentiable architecture search via perturbation-based regularization,'' in \emph{Int. Conf. Mach. Learn.}, vol. 119.\hskip 1em plus 0.5em minus 0.4em\relax PMLR, 13-18 Jul 2020, pp. 1554--1565.

\bibitem{zheng2021ad}
Z.~Zheng, L.~Yang, L.~Wang, and F.~Li, ``Ad-darts: Adaptive dropout for differentiable architecture search,'' in \emph{CAAI International Conference on Artificial Intelligence}.\hskip 1em plus 0.5em minus 0.4em\relax Springer, 2021, pp. 115--126.

\bibitem{xue2021idarts}
S.~Xue, R.~Wang, B.~Zhang, T.~Wang, G.~Guo, and D.~Doermann, ``Idarts: Interactive differentiable architecture search,'' in \emph{Int. Conf. Comput. Vis.}, 2021, pp. 1143--1152.

\bibitem{he2016deep}
K.~He, X.~Zhang, S.~Ren, and J.~Sun, ``Deep residual learning for image recognition,'' in \emph{IEEE Conf. Comput. Vis. Pattern Recog.}, 2016, pp. 1885--1894.

\bibitem{huang2019convolutional}
G.~Huang, Z.~Liu, G.~Pleiss, L.~Van Der~Maaten, and K.~Weinberger, ``Convolutional networks with dense connectivity,'' \emph{{IEEE} Trans. Pattern Anal. Mach. Intell.}, vol.~44, no.~12, pp. 8704--8716, 2019.

\bibitem{huang2018condensenet}
G.~Huang, S.~Liu, L.~Van~der Maaten, and K.~Q. Weinberger, ``Condensenet: An efficient densenet using learned group convolutions,'' in \emph{IEEE Conf. Comput. Vis. Pattern Recog.}, 2018, pp. 2752--2761.

\bibitem{yang2021condensenet}
L.~Yang, H.~Jiang, R.~Cai, Y.~Wang, S.~Song, G.~Huang, and Q.~Tian, ``Condensenet v2: Sparse feature reactivation for deep networks,'' in \emph{IEEE Conf. Comput. Vis. Pattern Recog.}, 2021, pp. 3569--3578.

\bibitem{wang2024towards}
J.~Wang, F.~Li, Y.~An, X.~Zhang, and H.~Sun, ``Toward robust lidar-camera fusion in bev space via mutual deformable attention and temporal aggregation,'' \emph{{IEEE} Trans. Circuits Syst. Video Technol.}, vol.~34, no.~7, pp. 5753--5764, 2024.

\bibitem{sun2022fast}
Q.~Sun, X.~Li, L.~Jiao, Y.~Ren, F.~Shang, and F.~Liu, ``Fast and effective: A novel sequential single-path search for mixed-precision-quantized networks,'' \emph{{IEEE} Trans. Cybern.}, 2022.

\bibitem{ji2022competitive}
J.~Ji, J.~Zhao, Q.~Lin, and K.~C. Tan, ``Competitive decomposition-based multiobjective architecture search for the dendritic neural model,'' \emph{{IEEE} Trans. Cybern.}, 2022.

\bibitem{li2022ds}
C.~Li, G.~Wang, B.~Wang, X.~Liang, Z.~Li, and X.~Chang, ``Ds-net++: Dynamic weight slicing for efficient inference in cnns and vision transformers,'' \emph{{IEEE} Trans. Pattern Anal. Mach. Intell.}, vol.~45, no.~4, pp. 4430--4446, 2022.

\bibitem{yang2024evolutionary}
S.~Yang, X.~Yu, Y.~Tian, X.~Yan, H.~Ma, and X.~Zhang, ``Evolutionary neural architecture search for transformer in knowledge tracing,'' \emph{Adv. Neural Inform. Process. Syst.}, vol.~36, 2024.

\bibitem{lyu2021multiobjective}
B.~Lyu, S.~Wen, K.~Shi, and T.~Huang, ``Multiobjective reinforcement learning-based neural architecture search for efficient portrait parsing,'' \emph{{IEEE} Trans. Cybern.}, 2021.

\bibitem{liu2021survey}
Y.~Liu, Y.~Sun, B.~Xue, M.~Zhang, G.~G. Yen, and K.~C. Tan, ``A survey on evolutionary neural architecture search,'' \emph{{IEEE} Trans. Neural Netw. Learn. Syst.}, vol.~34, no.~2, pp. 550--570, 2021.

\bibitem{dong2023diswot}
P.~Dong, L.~Li, and Z.~Wei, ``Diswot: Student architecture search for distillation without training,'' in \emph{IEEE Conf. Comput. Vis. Pattern Recog.}, 2023, pp. 11\,898--11\,908.

\bibitem{chen2024evoprompting}
A.~Chen, D.~Dohan, and D.~So, ``Evoprompting: Language models for code-level neural architecture search,'' \emph{Adv. Neural Inform. Process. Syst.}, vol.~36, 2023.

\bibitem{cai2020once}
H.~Cai, C.~Gan, T.~Wang, Z.~Zhang, and S.~Han, ``Once-for-all: Train one network and specialize it for efficient deployment,'' in \emph{Int. Conf. Learn. Represent.}, 2020.

\bibitem{guo2023pareto}
Y.~Guo, Y.~Chen, Y.~Zheng, Q.~Chen, P.~Zhao, J.~Huang, J.~Chen, and M.~Tan, ``Pareto-aware neural architecture generation for diverse computational budgets,'' in \emph{IEEE Conf. Comput. Vis. Pattern Recog.}, 2023, pp. 2247--2257.

\bibitem{lopes2024manas}
V.~Lopes, F.~M. Carlucci, P.~M. Esperan{\c{c}}a, M.~Singh, A.~Yang, V.~Gabillon, H.~Xu, Z.~Chen, and J.~Wang, ``Ma{NAS}: multi-agent neural architecture search,'' \emph{Machine Learning}, vol. 113, no.~1, pp. 73--96, 2024.

\bibitem{nasir2024llmatic}
M.~U. Nasir, S.~Earle, J.~Togelius, S.~James, and C.~Cleghorn, ``Llmatic: Neural architecture search via large language models and quality diversity optimization,'' in \emph{Proceedings of the Genetic and Evolutionary Computation Conference}, 2024, pp. 1110--1118.

\bibitem{heuillet2024efficient}
A.~Heuillet, A.~Nasser, H.~Arioui, and H.~Tabia, ``Efficient automation of neural network design: A survey on differentiable neural architecture search,'' \emph{ACM Computing Surveys}, vol.~56, no.~11, pp. 1--36, 2024.

\bibitem{xu2021partially}
Y.~Xu, L.~Xie, W.~Dai, X.~Zhang, X.~Chen, G.-J. Qi, H.~Xiong, and Q.~Tian, ``Partially-connected neural architecture search for reduced computational redundancy,'' \emph{{IEEE} Trans. Pattern Anal. Mach. Intell.}, vol.~43, no.~9, pp. 2953--2970, 2021.

\bibitem{dong2019searching}
X.~Dong and Y.~Yang, ``Searching for a robust neural architecture in four gpu hours,'' in \emph{IEEE Conf. Comput. Vis. Pattern Recog.}, 2019, pp. 1761--1770.

\bibitem{yan2021zeronas}
C.~Yan, X.~Chang, Z.~Li, W.~Guan, Z.~Ge, L.~Zhu, and Q.~Zheng, ``Zeronas: Differentiable generative adversarial networks search for zero-shot learning,'' \emph{{IEEE} Trans. Pattern Anal. Mach. Intell.}, vol.~44, no.~12, pp. 9733--9740, 2021.

\bibitem{zhang2021rs}
Z.~Zhang, S.~Liu, Y.~Zhang, and W.~Chen, ``Rs-darts: A convolutional neural architecture search for remote sensing image scene classification,'' \emph{Remote Sensing}, vol.~14, no.~1, p. 141, 2021.

\bibitem{shu2019understanding}
Y.~Shu, W.~Wang, and S.~Cai, ``Understanding architectures learnt by cell-based neural architecture search,'' \emph{Int. Conf. Learn. Represent.}, 2020.

\bibitem{chu2020fair}
X.~Chu, T.~Zhou, B.~Zhang, and J.~Li, ``Fair {DARTS}: Eliminating unfair advantages in differentiable architecture search,'' in \emph{Eur. Conf. Comput. Vis.}, 2020, pp. 465--480.

\bibitem{wang2021idarts}
H.~Wang, R.~Yang, D.~Huang, and Y.~Wang, ``idarts: Improving darts by node normalization and decorrelation discretization,'' \emph{{IEEE} Trans. Neural Netw. Learn. Syst.}, vol.~34, no.~4, pp. 1945--1957, 2021.

\bibitem{chu2021darts-}
X.~Chu, X.~Wang, B.~Zhang, S.~Lu, X.~Wei, and J.~Yan, ``Darts-: robustly stepping out of performance collapse without indicators,'' in \emph{Int. Conf. Learn. Represent.}, 2021.

\bibitem{xie2023architecture}
X.~Xie, Y.~Sun, Y.~Liu, M.~Zhang, and K.~C. Tan, ``Architecture augmentation for performance predictor via graph isomorphism,'' \emph{{IEEE} Trans. Cybern.}, 2023.

\bibitem{sun2019surrogate}
Y.~Sun, H.~Wang, B.~Xue, Y.~Jin, G.~G. Yen, and M.~Zhang, ``Surrogate-assisted evolutionary deep learning using an end-to-end random forest-based performance predictor,'' \emph{{IEEE} Trans. Evol. Comput.}, vol.~24, no.~2, pp. 350--364, 2019.

\bibitem{lee2018snip}
N.~Lee, T.~Ajanthan, and P.~H. Torr, ``Snip: Single-shot network pruning based on connection sensitivity,'' in \emph{Int. Conf. Learn. Represent.}, 2019.

\bibitem{wang2020picking}
C.~Wang, G.~Zhang, and R.~Grosse, ``Picking winning tickets before training by preserving gradient flow,'' in \emph{Int. Conf. Learn. Represent.}, 2020.

\bibitem{tanaka2020pruning}
H.~Tanaka, D.~Kunin, D.~L. Yamins, and S.~Ganguli, ``Pruning neural networks without any data by iteratively conserving synaptic flow,'' in \emph{Adv. Neural Inform. Process. Syst.}, vol.~33, 2020, pp. 6377--6389.

\bibitem{jorge2021progressive}
P.~D. Jorge, A.~Sanyal, H.~Behl, P.~Torr, G.~Rogez, and P.~K. Dokania, ``Progressive skeletonization: Trimming more fat from a network at initialization,'' in \emph{Int. Conf. Learn. Represent.}, 2021.

\bibitem{turner2020blockswap}
J.~Turner, E.~J. Crowley, M.~O'Boyle, A.~Storkey, and G.~Gray, ``Blockswap: Fisher-guided block substitution for network compression on a budget,'' in \emph{Int. Conf. Learn. Represent.}, 2020.

\bibitem{mellor2021neural}
J.~Mellor, J.~Turner, A.~Storkey, and E.~J. Crowley, ``Neural architecture search without training,'' in \emph{International Conference on Machine Learning (ICML)}, vol. 139, 2021, pp. 7588--7598.

\bibitem{abdelfattah2020zero}
M.~S. Abdelfattah, A.~Mehrotra, {\L}.~Dudziak, and N.~D. Lane, ``Zero-cost proxies for lightweight {NAS},'' in \emph{Int. Conf. Learn. Represent.}, 2020.

\bibitem{zhang2021differentiable}
M.~Zhang, S.~Su, S.~Pan, X.~Chang, W.~Huang, and G.~Haffari, ``Differentiable architecture search without training nor labels: A pruning perspective,'' \emph{arXiv preprint arXiv:2106.11542}, 2021.

\bibitem{lecun1989optimal}
Y.~LeCun, J.~Denker, and S.~Solla, ``Optimal brain damage,'' \emph{Adv. Neural Inform. Process. Syst.}, vol.~2, 1989.

\bibitem{molchanov2019importance}
P.~Molchanov, A.~Mallya, S.~Tyree, I.~Frosio, and J.~Kautz, ``Importance estimation for neural network pruning,'' in \emph{IEEE Conf. Comput. Vis. Pattern Recog.}, 2019, pp. 11\,264--11\,272.

\bibitem{molchanov2016pruning}
P.~Molchanov, S.~Tyree, T.~Karras, T.~Aila, and J.~Kautz, ``Pruning convolutional neural networks for resource efficient inference,'' in \emph{Int. Conf. Learn. Represent.}, 2017.

\bibitem{greff2016highway}
K.~Greff, R.~K. Srivastava, and J.~Schmidhuber, ``Highway and residual networks learn unrolled iterative estimation,'' in \emph{Int. Conf. Learn. Represent.}, 2016.

\bibitem{sun2023simple}
M.~Sun, Z.~Liu, A.~Bair, and J.~Z. Kolter, ``A simple and effective pruning approach for large language models,'' \emph{arXiv preprint arXiv:2306.11695}, 2023.

\bibitem{hu2020dsnas}
S.~Hu, S.~Xie, H.~Zheng, C.~Liu, J.~Shi, X.~Liu, and D.~Lin, ``Dsnas: Direct neural architecture search without parameter retraining,'' in \emph{IEEE Conf. Comput. Vis. Pattern Recog.}, 2020, pp. 12\,084--12\,092.

\bibitem{xie2019snas}
S.~Xie, H.~Zheng, C.~Liu, and L.~Lin, ``{SNAS}: stochastic neural architecture search,'' in \emph{Int. Conf. Learn. Represent.}, 2019.

\bibitem{zhang2021idarts}
M.~Zhang, S.~W. Su, S.~Pan, X.~Chang, E.~M. Abbasnejad, and R.~Haffari, ``idarts: Differentiable architecture search with stochastic implicit gradients,'' in \emph{Int. Conf. Mach. Learn.}, 2021, pp. 12\,557--12\,566.

\bibitem{qian2022meets}
G.~Qian, X.~Zhang, G.~Li, C.~Zhao, Y.~Chen, X.~Zhang, B.~Ghanem, and J.~Sun, ``When nas meets trees: An efficient algorithm for neural architecture search,'' in \emph{IEEE Conf. Comput. Vis. Pattern Recog.}, 2022.

\bibitem{dong2020bench}
X.~Dong and Y.~Yang, ``Nas-bench-201: Extending the scope of reproducible neural architecture search,'' in \emph{Int. Conf. Learn. Represent.}, 2020.

\bibitem{krizhevsky2009learning}
A.~Krizhevsky, ``Learning multiple layers of features from tiny images,'' \emph{Master's thesis, University of Tront}, 2009.

\bibitem{chrabaszcz2017downsampled}
P.~Chrabaszcz, I.~Loshchilov, and F.~Hutter, ``A downsampled variant of imagenet as an alternative to the cifar datasets,'' \emph{arXiv preprint arXiv:1707.08819}, 2017.

\bibitem{chen2019pdarts}
X.~Chen, L.~Xie, J.~Wu, and Q.~Tian, ``Progressive differentiable architecture search: Bridging the depth gap between search and evaluation,'' in \emph{Int. Conf. Comput. Vis.}, 2019, pp. 1294--1303.

\bibitem{xue2024self}
Y.~Xue, X.~Han, and Z.~Wang, ``Self-adaptive weight based on dual-attention for differentiable neural architecture search,'' \emph{{IEEE} Trans. Ind. Informat.}, vol.~20, no.~4, pp. 6394--6403, 2024.

\bibitem{xiao2022shapley}
H.~Xiao, Z.~Wang, Z.~Zhu, J.~Zhou, and J.~Lu, ``Shapley-nas: discovering operation contribution for neural architecture search,'' in \emph{IEEE Conf. Comput. Vis. Pattern Recog.}, 2022.

\bibitem{chen2020drnas}
X.~Chen, R.~Wang, M.~Cheng, X.~Tang, and C.-J. Hsieh, ``Drnas: Dirichlet neural architecture search,'' \emph{arXiv preprint arXiv:2006.10355}, 2020.

\bibitem{deng2009imagenet}
J.~Deng, W.~Dong, R.~Socher, L.-J. Li, K.~Li, and L.~Fei-Fei, ``Imagenet: A large-scale hierarchical image database,'' in \emph{IEEE Conf. Comput. Vis. Pattern Recog.}\hskip 1em plus 0.5em minus 0.4em\relax Ieee, 2009, pp. 248--255.

\bibitem{ma2018shufflenet}
N.~Ma, X.~Zhang, H.-T. Zheng, and J.~Sun, ``Shufflenet v2: Practical guidelines for efficient cnn architecture design,'' in \emph{Eur. Conf. Comput. Vis.}, 2018, pp. 116--131.

\bibitem{cubuk2020randaugment}
E.~D. Cubuk, B.~Zoph, J.~Shlens, and Q.~V. Le, ``Randaugment: Practical automated data augmentation with a reduced search space,'' in \emph{Proceedings of the IEEE/CVF conference on computer vision and pattern recognition workshops}, 2020, pp. 702--703.

\bibitem{zhang2021neural}
X.~Zhang, P.~Hou, X.~Zhang, and J.~Sun, ``Neural architecture search with random labels,'' in \emph{IEEE Conf. Comput. Vis. Pattern Recog.}, 2021, pp. 10\,907--10\,916.

\bibitem{liu2020labels}
C.~Liu, P.~Doll{\'a}r, K.~He, R.~Girshick, A.~Yuille, and S.~Xie, ``Are labels necessary for neural architecture search?'' in \emph{Eur. Conf. Comput. Vis.}\hskip 1em plus 0.5em minus 0.4em\relax Springer, 2020, pp. 798--813.

\bibitem{lin2017focal}
T.-Y. Lin, P.~Goyal, R.~Girshick, K.~He, and P.~Doll{\'a}r, ``Focal loss for dense object detection,'' in \emph{Int. Conf. Comput. Vis.}, 2017, pp. 2980--2988.

\bibitem{lin2014microsoft}
T.-Y. Lin, M.~Maire, S.~Belongie, J.~Hays, P.~Perona, D.~Ramanan, P.~Doll{\'a}r, and C.~L. Zitnick, ``Microsoft coco: Common objects in context,'' in \emph{Eur. Conf. Comput. Vis.}\hskip 1em plus 0.5em minus 0.4em\relax Springer, 2014, pp. 740--755.

\bibitem{lee2019signal}
N.~Lee, T.~Ajanthan, S.~Gould, and P.~H. Torr, ``A signal propagation perspective for pruning neural networks at initialization,'' in \emph{Int. Conf. Learn. Represent.}, 2019.

\end{thebibliography}

}

\begin{IEEEbiography}[{\includegraphics[width=1in,height=1.25in,clip,keepaspectratio]{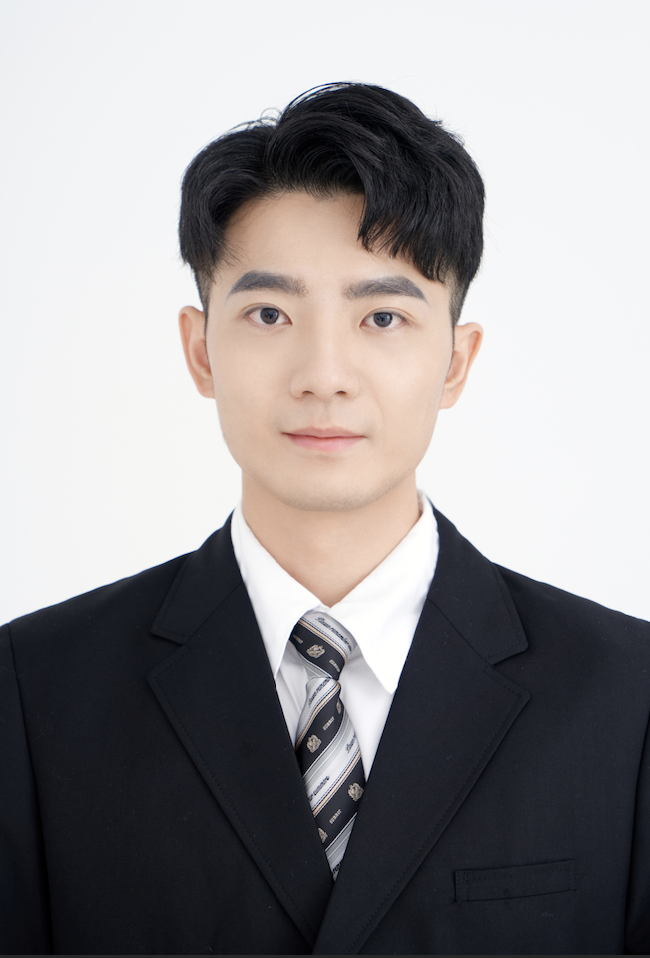}}]{Le Yang} received his B.S. degree in Department of Automation, Northwestern Polytechnical University in 2015, and the Ph.D. degree from the Department of Automation, Tsinghua University in 2021. He is currently an assistant professor with the School of Information and Communications Engineering, Xi’an Jiaotong University. His main research interests include deep learning and computer vision.
\end{IEEEbiography}

\begin{IEEEbiography}[{\includegraphics[width=1in,height=1.25in,clip,keepaspectratio]{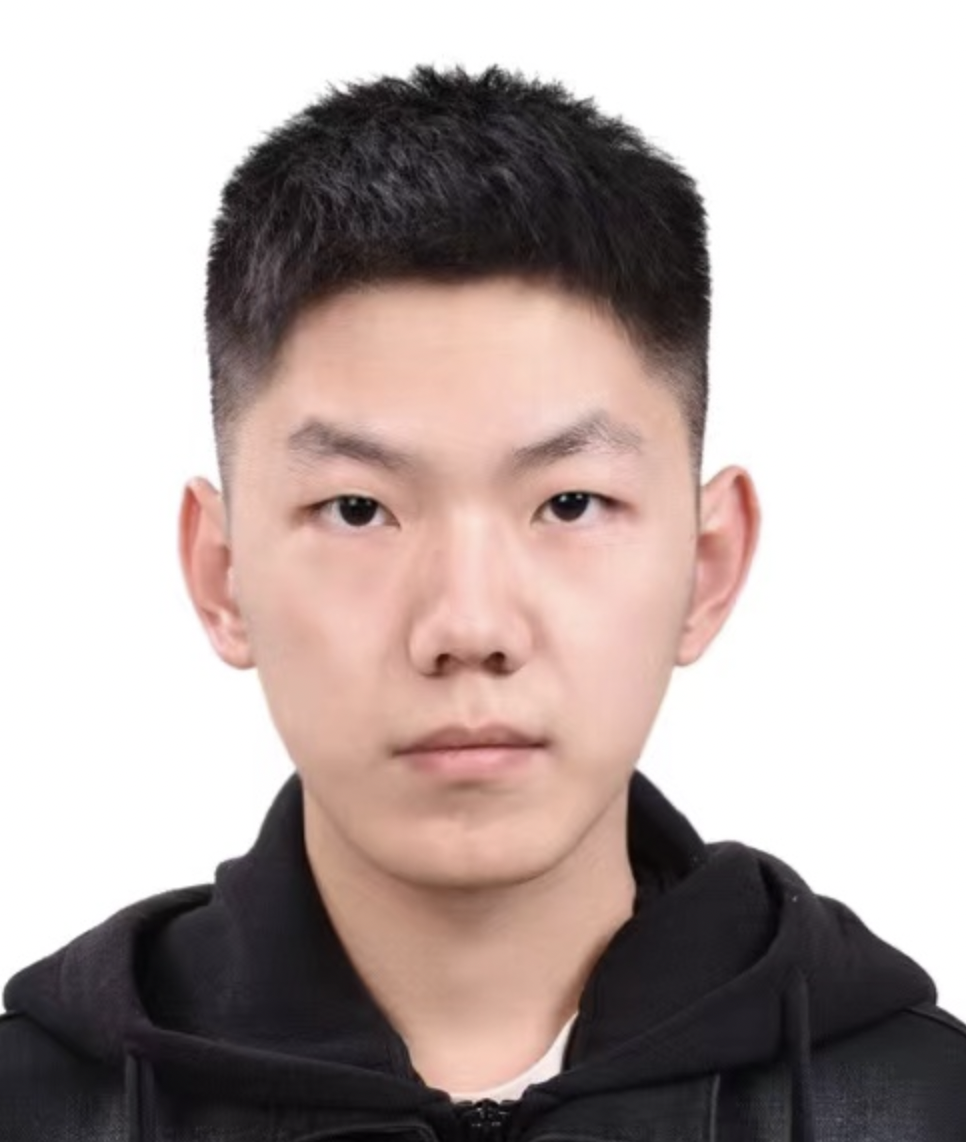}}]{Ziwei Zheng} received the B.S. degree in information engineering from Xi’an Jiaotong University, Xi’an, China, in 2021. From 2021, he started his M.S. degree at the School of Information and Communications Engineering, Xi’an Jiaotong University. His current research interests include computer vision, deep learning, and neural architecture search.
\end{IEEEbiography}

\begin{IEEEbiography}[{\includegraphics[width=1in,height=1.25in,clip,keepaspectratio]{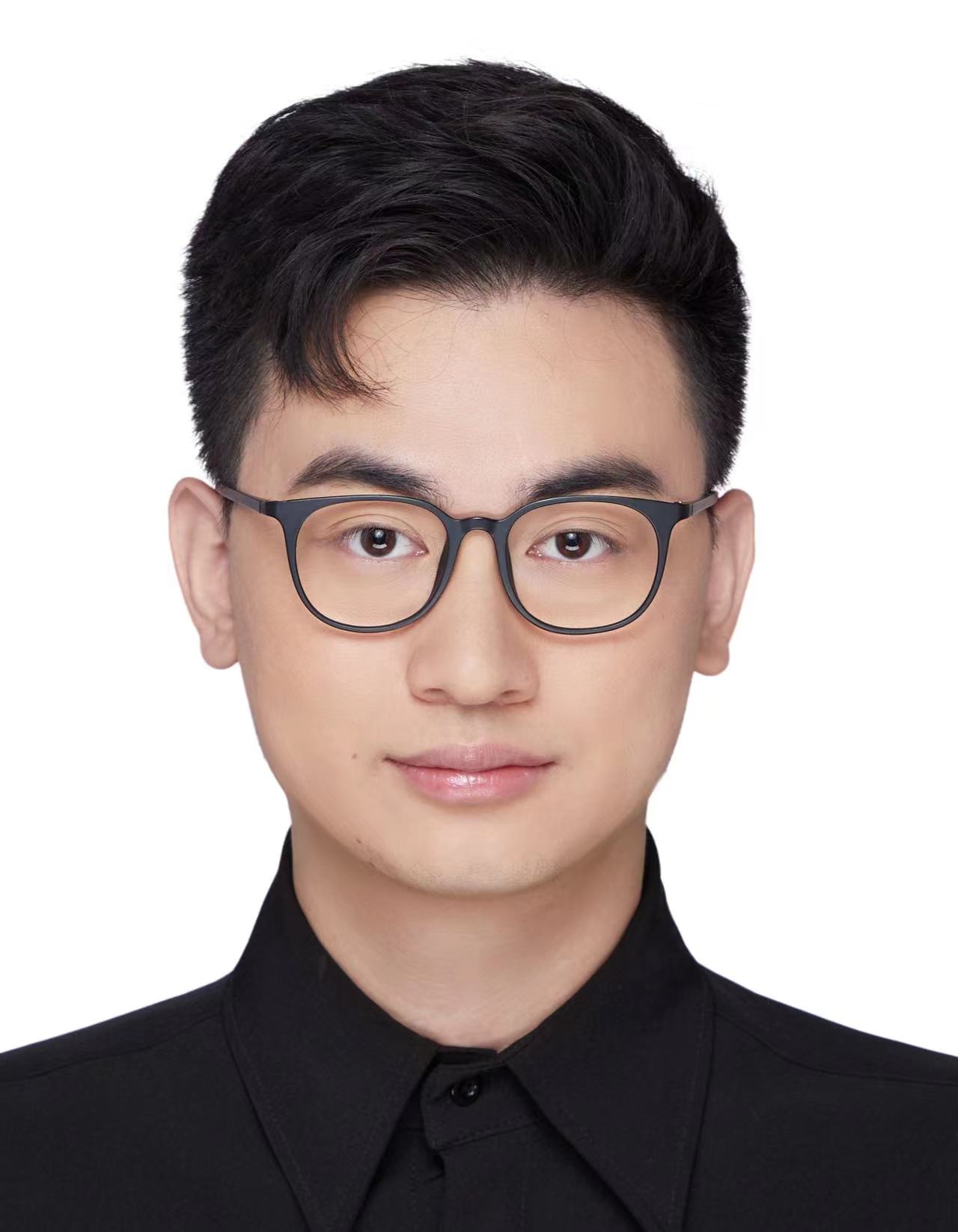}}]{Yizeng Han} received the B.S degree and the Ph.D degree from the department of Automation, Tsinghua University, China in 2018 and 2024, respectively. He is currently a researcher at Damo Academy, Alibaba group. His research interests include computer vision and deep learning, especially in dynamic neural networks and efficient AI.
\end{IEEEbiography}

\begin{IEEEbiography}[{\includegraphics[width=1in,height=1.25in,clip,keepaspectratio]{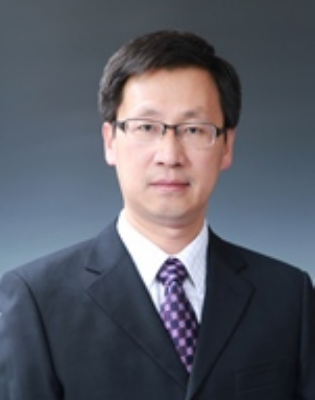}}]{Shiji Song} (Senior Member, IEEE) received the PhD degree in mathematics from the Department of Mathematics, Harbin Institute of Technology, Harbin, China, in 1996. He is currently a professor with the Department of Automation, Tsinghua University, Beijing, China. His current research interests include pattern recognition, system modeling, and optimization and control.
\end{IEEEbiography}

\begin{IEEEbiography}[{\includegraphics[width=1in,height=1.25in,clip,keepaspectratio]{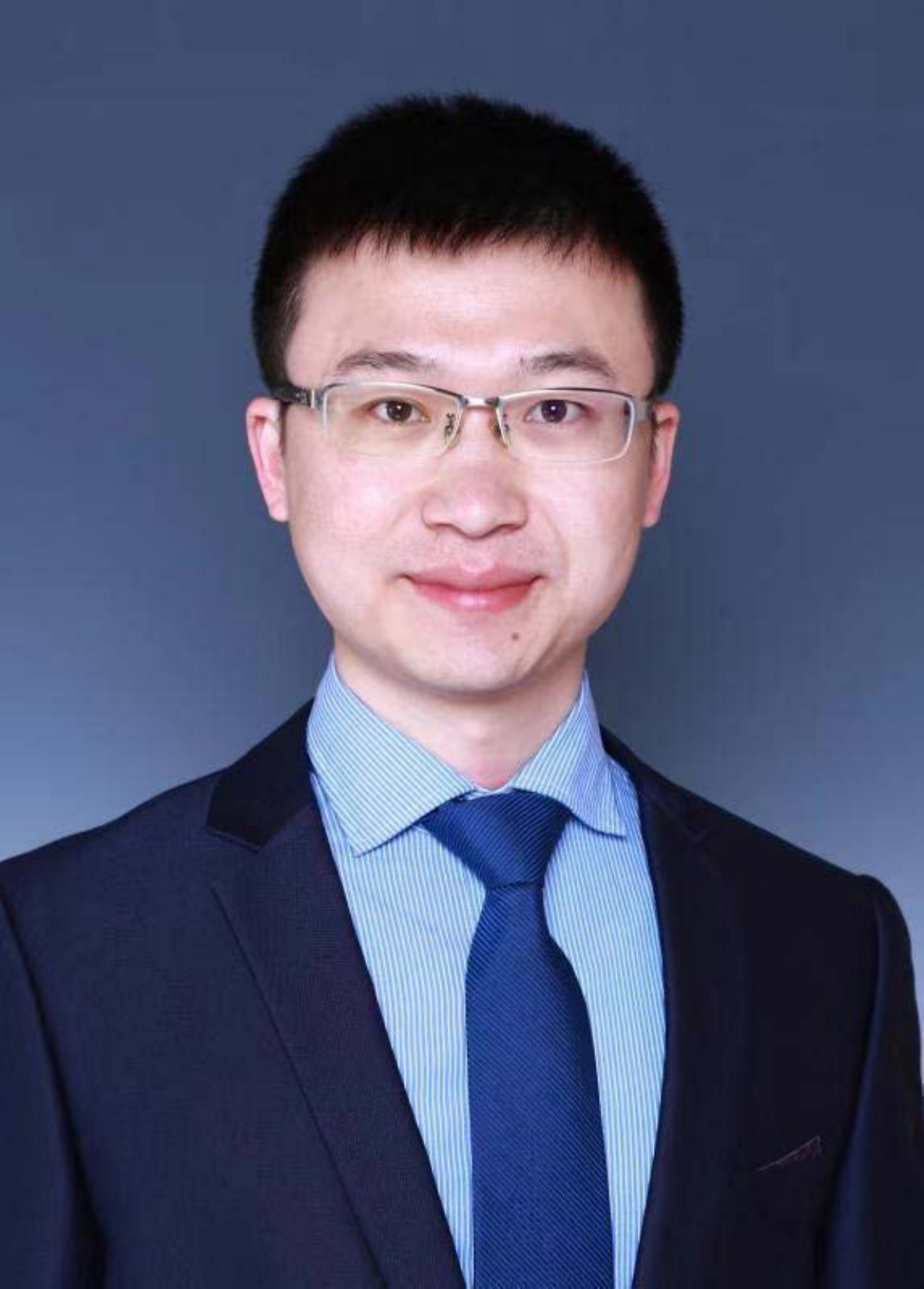}}]{Gao Huang} (Member, IEEE) received the B.S. degree from the School of Automation Science and Electrical Engineering, Beihang University, Beijing, China, in 2009, and the Ph.D. degree from the Department of Automation, Tsinghua University, Beijing, in 2015. He was was a Post-Doctoral Researcher with Department of Computer Science, Cornell University from 2015 to 2018. He is currently an associate professor at the Department of Automation, Tsinghua University. His research interests include machine learning and computer vision.
\end{IEEEbiography}

\begin{IEEEbiography}[{\includegraphics[width=1in,height=1.25in,clip,keepaspectratio]{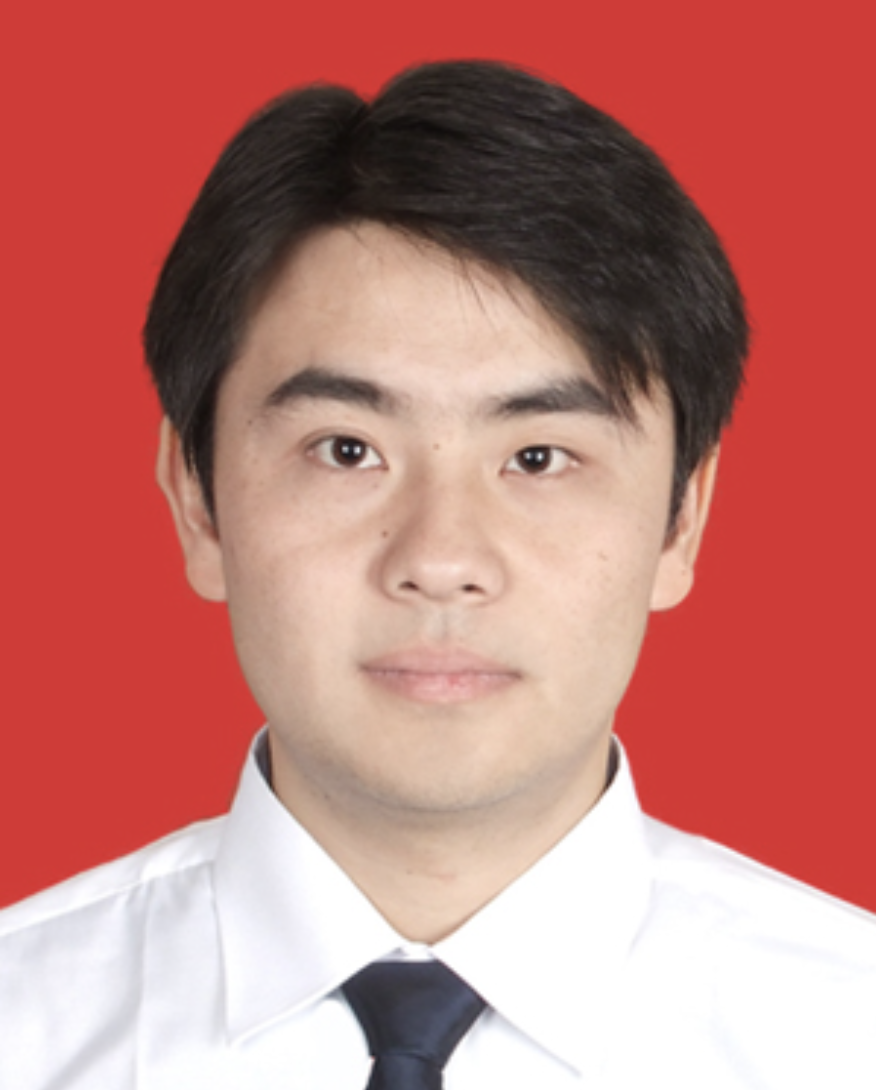}}]{Fan Li} (Senior Member, IEEE) received the B.S. and Ph.D. degrees in information engineering from Xi’an Jiaotong University, Xi’an, China, in 2003 and 2010, respectively. From 2017 to 2018, he was a Visiting Scholar with the Department of Electrical and Computer Engineering, University of California at San Diego. He is currently a Professor with the School of Information and Communications Engineering, Xi’an Jiaotong University. His research interests include multimedia communication, image/video coding, and artificial intelligence. 
\end{IEEEbiography}

\end{document}